%% file: main.tex
\documentclass[11pt]{article}
\usepackage[margin=1in]{geometry}

\usepackage{times}
\usepackage{natbib}

\usepackage[utf8]{inputenc} 
\usepackage[T1]{fontenc}    
\usepackage{hyperref}       
\usepackage{url}            
\usepackage{booktabs}       
\usepackage{amsfonts}       
\usepackage{nicefrac}       
\usepackage{microtype}      
\usepackage{xcolor}         

\usepackage{graphicx}
\usepackage{multirow}
\usepackage{pifont}
\usepackage{xspace}
\usepackage{upgreek} 
\usepackage{tikz}
\usepackage{subcaption}

\usepackage{algorithm}
\usepackage{algpseudocode}
\usepackage{amsmath}
\usepackage[capitalise]{cleveref}

\usepackage{wrapfig}
\usepackage{lipsum}

\input{math_commands.tex}

\def\BibTeX{{\rm B\kern-.05em{\sc i\kern-.025em b}\kern-.08em
    T\kern-.1667em\lower.7ex\hbox{E}\kern-.125emX}}

\title{Conflict-Aware Adversarial Training}

\author{%
  Zhiyu Xue\textsuperscript{1} \hspace{0.2cm} ~Haohan Wang\textsuperscript{2} \hspace{0.2cm} ~Yao Qin\textsuperscript{1} \hspace{0.2cm} ~Ramtin Pedarsani\textsuperscript{1}\\
  \textsuperscript{1}UC Santa Barbara, \textsuperscript{2}University of Illinois Urbana-Champaign\\
  \texttt{\{zhiyuxue,yaoqin,ramtin\}@ucsb.edu, haohanw@illinois.edu}
}
\date{}

\begin{document}

\maketitle

\begin{abstract}
  Adversarial training is the most effective method to obtain adversarial robustness for deep neural networks by directly involving adversarial samples in the training procedure. To obtain an accurate and robust model, the weighted-average method is applied to optimize standard loss and adversarial loss simultaneously. In this paper, we argue that the weighted-average method does not provide the best tradeoff for the standard performance and adversarial robustness. We argue that the failure of the weighted-average method is due to the conflict between the gradients derived from standard and adversarial loss, and further demonstrate such a conflict increases with attack budget theoretically and practically. To alleviate this problem, we propose a new trade-off paradigm for adversarial training with a conflict-aware factor for the convex combination of standard and adversarial loss, named \textbf{Conflict-Aware Adversarial Training~(CA-AT)}. Comprehensive experimental results show that CA-AT consistently offers a superior trade-off between standard performance and adversarial robustness under the settings of adversarial training from scratch and parameter-efficient finetuning. 
\end{abstract}

\input{Sections/intro}

\input{Sections/related_works}
\input{Sections/Con}
\input{Sections/Method}
\input{Sections/exp}
\section{Conclusion \& Outlook}
\label{sec:Conclusion}
In this work, we illustrate that the weighted-average method in AT is not capable of achieving the `near-optimal' trade-off between standard and adversarial accuracy due to the gradient conflict existing in the training process. We demonstrate the existence of such a gradient conflict and its relation to the attack budget of adversarial samples used in AT practically and theoretically. Based on this phenomenon, we propose a new trade-off framework for AT called Conflict-Aware Adversarial Training (CA-AT) to alleviate the conflict by gradient operation, as well as applying a conflict-aware trade-off factor to the convex combination of standard and adversarial loss functions. Extensive results demonstrate the effectiveness of CA-AT for gaining trade-off results under the setting of training from scratch and PEFT. 

For future work, we plan to undertake a more detailed exploration of the gradient conflict phenomenon in AT from the data-centric perspective. We hold the assumption that some training samples can cause serious gradient conflict, while others do not. We will evaluate this assumption in the future work, and intend to reveal the influence of training samples causing gradient conflict.

\clearpage
\bibliography{ref}
\bibliographystyle{plainnat}

\clearpage
\appendix
\input{Sections/appendix}

\end{document}

%% file: math_commands.tex
\usepackage{amsmath,amsfonts,bm}

\def\Ladv{{\mathcal{L}_{\text{a}}}}
\def\Lc{{\mathcal{L}_{\text{c}}}}
\def\gc{g_{\text{c}}}
\def\ga{g_{\text{a}}}

\def\eqref#1{equation~\ref{#1}}

\def\1{\bm{1}}

\DeclareMathAlphabet{\mathsfit}{\encodingdefault}{\sfdefault}{m}{sl}
\SetMathAlphabet{\mathsfit}{bold}{\encodingdefault}{\sfdefault}{bx}{n}

\def\sE{{\mathbb{E}}}



%% file: Sections/intro.tex

\section{Introduction}
\label{sec:intro}


Deep learning models have achieved exemplary performance across diverse application domains~\citep{he2017mask,vaswani2017attention,ouyang2022training,rombach2022high,radford2021learning}. However, they remain vulnerable to adversarial samples produced by adversarial attacks~\citep{goodfellow2014explaining,liu2016delving,moosavi2016deepfool}. Deep learning models can easily be fooled into making mistakes by adding an imperceptible noise produced by adversarial attacks to the standard sample. To solve this problem, many methods have been proposed to improve the robustness against adversarial samples~\citep{cai2018curriculum,chakraborty2018attack,madry2018towards}, among which \textbf{adversarial training~(AT)} has been proven to be the most effective strategy~\citep{madry2018towards,athalye2018obfuscated,qian2022survey,bai2021recent}. Specifically, AT aims to enhance model robustness by directly involving adversarial samples during training. They used adversarial examples to construct the adversarial loss functions for parameter optimization. The adversarial loss can be formulated as a min-max optimization objective, where the adversarial samples are generated by the inner maximization, and the model parameters are optimized by the outer minimization to reduce the empirical risk for adversarial samples.

The trade-off between standard and adversarial accuracy is a key factor for the real-world applications of AT~\citep{tsipras2018robustness,balaji2019instance,yang2020closer,stutz2019disentangling,zhang2019theoretically}. Although AT can improve robustness against adversarial samples, it also undermines the performance on standard samples. Existing AT methods~\citep{madry2018towards,cai2018curriculum,zhang2019theoretically,wang2019improving} design a hybrid loss by combining standard loss and an adversarial loss linearly, where the linear coefficient typically serves as the trade-off factor.

In this paper, we argue that linearly weighted-average method for AT, as well as the \textbf{Vanilla AT}, cannot achieve a `near-optimal' trade-off. In other words, it fails to approximately achieve the Pareto optimal points on the Pareto front of standard and adversarial accuracies. We find that the conflict between the parameter gradient derived from standard loss~(\textbf{standard gradient}) and the one derived from adversarial loss~(\textbf{adversarial gradient}) is the main source of this failure. Such a gradient conflict causes the model parameter to be stuck in undesirable local optimal points, and it becomes more severe with the increase of adversarial attack budget. In addition, to obtain adversarial robustness, linearly weighted-average method usually sacrifices too much performance on standard samples, which hinders AT from real-world applications.

To solve the problems mentioned above, we propose \textbf{Conflict-Aware Adversarial Training~(CA-AT)} to mitigate the conflict during adversarial training. Inspired by gradient surgery~\citep{yu2020gradient} in multi-task learning, CA-AT utilizes a new trade-off factor defined as the angle between the standard and adversarial gradients. As depicted in \cref{fig:arch}, if the angle is larger than the pre-defined trade-off factor $\gamma$,  CA-AT will project the adversarial gradient onto the `cone' around the standard gradient constructed based on the pre-defined trade-off factor; otherwise, it will use the standard gradient to optimize the model parameter $\theta$ directly. Compared to the linearly weighted-average AT with a fixed trade-off factor, CA-AT can boost both standard and adversarial accuracy. Our primary contributions are summarized as follows:

\begin{figure}[t]
    \centering
    \includegraphics[width=\textwidth]{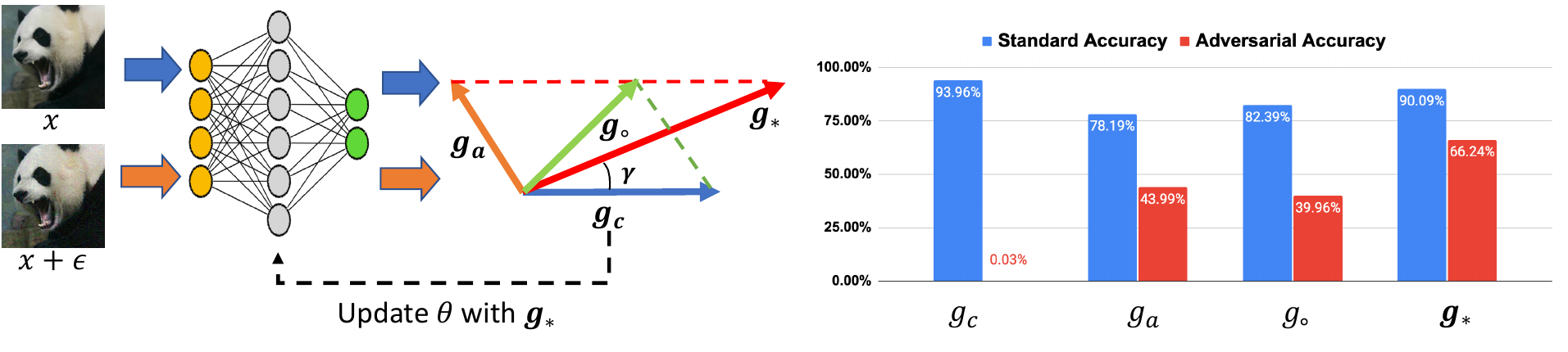}
    \caption{The key motivation of CA-AT aims to solve the conflict between clean gradient $g_{c}$ and adversarial gradient $\ga$. Unlike the existing weighted-averaged method optimizing model parameter $\theta$ by $g_{\circ}$ as the average of $\gc$ and $\ga$~(Vanilla AT), CA-AT utilizes $g_{*}$ for parameter optimization by gradient projection based on a new trade-off factor $\phi$. The bar chart on the right side illustrates that the model optimized by $g_{*}$~(highlighted as the boldface) can achieve better standard accuracy~(blue bar) and adversarial accuracy~(red bar) compared to models optimized by $g_{\circ}$. 
    The results of the bar chart on the right are produced by training a ResNet18 on CIFAR10 against the PGD~\citep{madry2017PGD} attack.}
    \label{fig:arch}
\end{figure}
\begin{enumerate}
    \item We shed light on the existence of conflict between standard and adversarial gradient which causes a sub-optimal trade-off between standard and adversarial accuracy in AT, when we optimize standard and adversarial loss in weighted-average paradigm by a fixed trade-off factor.
    \item To alleviate the gradient conflict in AT, we propose a new paradigm called Conflict-Aware Adversarial Training~(CA-AT). It achieves a better trade-off between standard and adversarial accuracy compared to Vanilla AT. 
    \item Through comprehensive experiments across a wide range of settings, we demonstrate CA-AT consistently improves the trade-off between standard and adversarial accuracy in the context of training from scratch and parameter-efficient finetuning~(PEFT), across diverse adversarial loss functions, adversarial attack types, model architectures, and datasets.
\end{enumerate}

%% file: Sections/related_works.tex
\section{Related Works}
\label{Sec: Related}
\textbf{Adversarial Training.}
Adversarial training~(AT) is now broadly considered as the most effective method to achieve adversarial robustness for deep learning models~\citep{qian2022survey,singh2024revisitingAT}. The key idea of AT is to involve adversarial samples during the training process. Existing works for AT can be mainly grouped into regularization-driven and strategy-driven. For regularization-driven AT methods, the goal is to design an appropriate loss function for adversarial samples, such as cross-entropy~\citep{madry2017PGD}, logits pairing~(CLP)~\citep{kannan2018logitpair}, and TRADES~\citep{zhang2019theoretically}. On the other hand, strategy-driven AT methods focus on improving adversarial robustness by designing appropriate training strategies. For example, ensemble AT~\citep{tramer2017ensemble,yang2020ensemble} alleviates the sharp parameter curvature by utilizing adversarial examples generated from different target models, curriculum AT~\citep{cai2018curriculum} gains adversarial robustness progressively by learning from easy adversarial samples to hard adversarial samples, and adaptive AT~\citep{ding2018mma,cheng2020cat,jia2022learnable} improves adversarial robustness by adjusting the attack intensity and attack methods. With the development of large-scale pretrained models~\citep{kolesnikov2020bit,dong2020adverpretrain}, \citep{jia2024improving,hua2023initialization} demonstrates the superiority of adversarial PEFT of robust pretrained models, compared to adversarial training from scratch.  

However, strategy-based AT methods need to involve additional attack methods or target models in the training process, which will increase the time and space complexity when we apply them. CA-AT can improve both standard and adversarial performance without any increasing cost of training time and computing resources. Besides, CA-AT can also work well with different adversarial loss functions including CLP and TRADES, which will be shown in the section of experimental results. Besides, we find out that CA-AT can work well under the setting of adversarial PEFT.

\textbf{Gradients Operation.}
Gradients Operation, also known as gradient surgery~\citep{yu2020gradient}, aims to improve model performance by directly operating the parameter gradient during training. It was first presented in the area of multi-task learning to alleviate the gradient conflict between loss functions designed for different tasks. The conflict can be measured by cosine similarity~\citep{yu2020gradient} or Euclidean distance~\citep{liu2021conflict} between the gradients derived from different loss functions. Besides, multi-task learning, \citep{mansilla2021domain} incorporates gradient operation to encourage gradient agreement among different source domains, enhancing the model's generalization ability to the unseen domain, and \citep{chaudhry2018agem,yang2023restricted} alleviate the forgetting issue in continual learning by projecting the gradients from the current task to the orthogonal direction of gradients derived from the previous task.

We are the first work to observe the gradient conflict between standard and adversarial loss during AT and further reveal its relation to adversarial attack budget. Moreover, we propose a new trade-off paradigm specifically designed for AT based on gradient operation. It can achieve a better trade-off compared to Vanilla AT and guarantee the standard performance well. 

%% file: Sections/Con.tex
\section{Gradient Conflict in AT}
\label{sec:Con}

In this section, we will discuss the occurrence of gradient conflict in AT via a synthetic dataset and real-world datasets such as CIFAR10 and CIFAR100. Additionally, we demonstrate such a conflict will become more serious with the increase of the attack budget theoretically and practically.
\subsection{Preliminaries \& Notations}
Considering a set of images, each image $x\in \mathbb{R}^{d}$ and its label $y\in \mathbb{R}^{l}$ is drawn i.i.d. from distribution $\mathcal{D}$. The classifier $f: \mathbb{R}^{d} \rightarrow \mathbb{R}^{l}$ parameterized by $\theta$ aims to map an input image to the probabilities of the classification task. The objective of AT is to ensure that $f$ does not only perform well on $x$, but also manifests robustness against adversarial perturbation $\epsilon$ bounded by attack budget $\delta$ as $\|\epsilon\|_{p} \leq \delta$, where $p$ determinates the $L_{p}$ norm constraint on the perturbations $\epsilon$ commonly taking on the values of $\infty$ or $2$. The perturbation $\epsilon$ can be defined as $\epsilon = \arg\max_{\|\epsilon\|_{p}\leq \delta}\mathcal{L}(x+\epsilon,y;\theta)$, which can be approximated by gradient-based adversarial attacks such as PGD. Throughout the remaining part of this paper, we refer to $x$ as the standard sample and $x+\epsilon$ as the adversarial sample.

We define clean loss $\Lc = \mathcal{L}(x,y;\theta)$ and adversarial loss $\Ladv = \mathcal{L}(x+\epsilon,y;\theta)$, respectively. $\mathcal{L}$ is the loss function for classification task~(e.g. cross-entropy). As shown in \cref{Eq: AdverT}, the goal of adversarial training is to obtain the parameter $\theta$ that can be both accurate and robust. 
\begin{align}
    \min_{\theta}(\mathbb{E}_{(x,y) \sim \mathcal{D}}[\Lc], \mathbb{E}_{(x,y) \sim \mathcal{D}}[\Ladv]) \label{Eq: AdverT}
\end{align}
For vanilla AT, as mentioned in \cref{Sec: Related}, optimizing a hybrid loss containing standard loss $\Lc$ and adversarial loss $\Ladv$ is a widely-used method for solving \cref{Eq: AdverT}. As shown in \cref{Eq: AdverT_Linear}, existing works~\citep{wang2019improving,zhang2019theoretically,kannan2018logitpair} construct such a hybrid loss by using a linear-weighted approach for $\Lc$ and $\Ladv$.
\begin{gather}
\label{Eq: AdverT_Linear}
\min_{\theta} \mathbb{E}_{(x,y) \sim \mathcal{D}} [\lambda\Ladv+(1-\lambda)\Lc],
\end{gather}
where $\lambda \in [0,1]$ serves as a fixed hyper-parameter for the trade-off between $\Lc$ and $\Ladv$. Refer to \cref{fig:arch}, the optimization process of \cref{Eq: AdverT_Linear} can be described as utilizing $g_{\circ} = (1-\lambda)g_{c} + \lambda g_{a}$ to update $\theta$ at each optimization step, where $\gc = \frac{\partial\Lc}{\partial \theta}$ and $\ga = \frac{\partial \Ladv}{\partial \theta}$ represent standard and adversarial gradients, respectively. 

To measure how well we can solve \cref{Eq: AdverT}, we define a metric $\mu = ||\gc||_{2} \cdot ||\ga||_{2} \cdot (1-\cos(\gc,\ga))$. The basic motivation for the consideration of $\mu$ is that it should combine three kinds of signals during AT simultaneously: \textbf{(1)} $||\gc||_{2}$ reflects the convergence of clean loss $\Lc$, \textbf{(2)} $||\ga||_{2}$ reflects the convergence of adversarial loss $\Ladv$, and \textbf{(3)} $(1-\cos(\gc,\ga))$ reflects the directional conflict between $\gc$ and $\ga$.
Based on \textbf{(1)}, \textbf{(2)}, and \textbf{(3)}, a small $\mu$ implies that both $\Lc$ and $\Ladv$ have converged well while reaching a consensus on the optimization direction for the next step.
\subsection{Theoretical \& Experimental Support for Motivation}
We introduce Theorem 1 that demonstrates $\mu$ can be bounded by the input dimension $d$ and perturbation budget $\delta$ in AT. \\ \quad \\
\textbf{Theorem 1.} {\it Consider the gradient conflict $\mu = ||\gc||_{2} \cdot ||\ga||_{2} \cdot (1-\cos(\gc,\ga))$ and suppose that the input $x$ is a $d$-dimensional vector.
\begin{enumerate}
    \item Given the $L_{2}$ restriction for $\epsilon$ as $||\epsilon||_{2} \leq \delta$, we have $\mu \leq \mathcal{O}(\delta^{2})$.
    \item Given the $L_{\infty}$ restriction for $\epsilon$ as $||\epsilon||_{\infty} \leq \delta$, we have $\mu \leq \mathcal{O}(d^{2}\delta^{2})$.
\end{enumerate}}
The intuitive understanding of Theorem 1 is that with the increasing attack budget $\delta$, the adversarial samples in AT will move further from the distribution $\mathcal{D}$ of standard samples. The conflict between $g_{a}$ and $g_{c}$ will become more serious, and $L_{a}$ and $L_{c}$ will be hard to converge. Therefore, the upper bound of $\mu$ will become larger. The proof of Theorem 1 will be shown in the appendix.

\textbf{Synthetic Experiment.} In order to show the implications of Theorem 1 empirically, we introduce the synthetic experiment as a binary classification task by selecting digit one and digit two from MNIST with a resolution of $32\times 32$, and train a logistic regression model parameterized by $w\in\mathbb{R}^{(32\times 32) \times 2}$ via BCE loss by vanilla AT for 20 epochs, where $\epsilon$ is contained by its $L_{\infty}$ norm as $\|\epsilon\|_{\infty}\leq \delta$, and $\lambda=0.5$ serves as the trade-off factor between standard and adversarial loss.
\begin{figure}[t]
  \centering
  \begin{subfigure}[b]{0.5\textwidth}
    \centering
    \includegraphics[width=\textwidth]{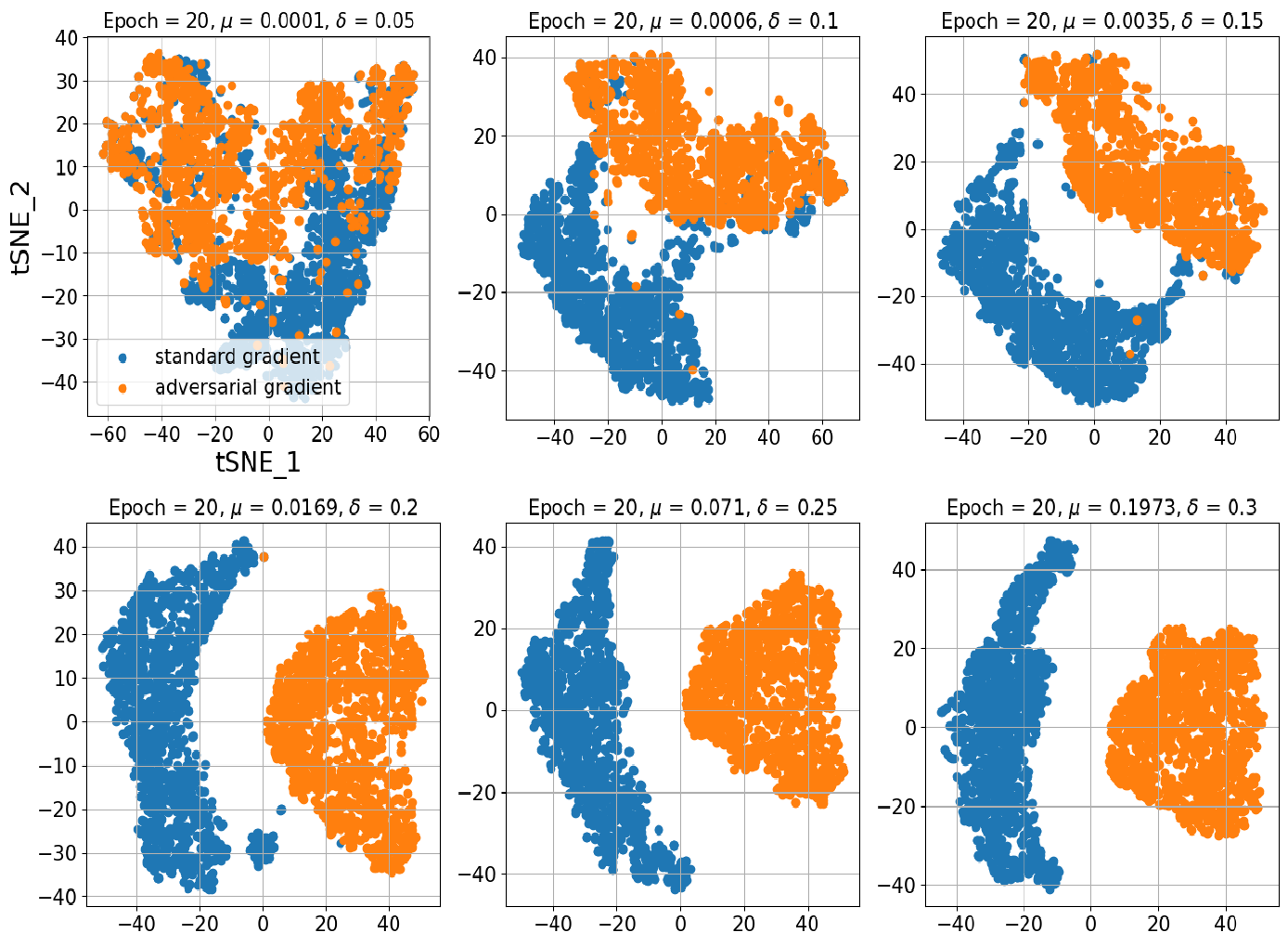}
    \caption{}
    \label{fig:Toy(a)}
  \end{subfigure}
  \begin{subfigure}[b]{0.45\textwidth}
    \centering
    \includegraphics[width=\textwidth]{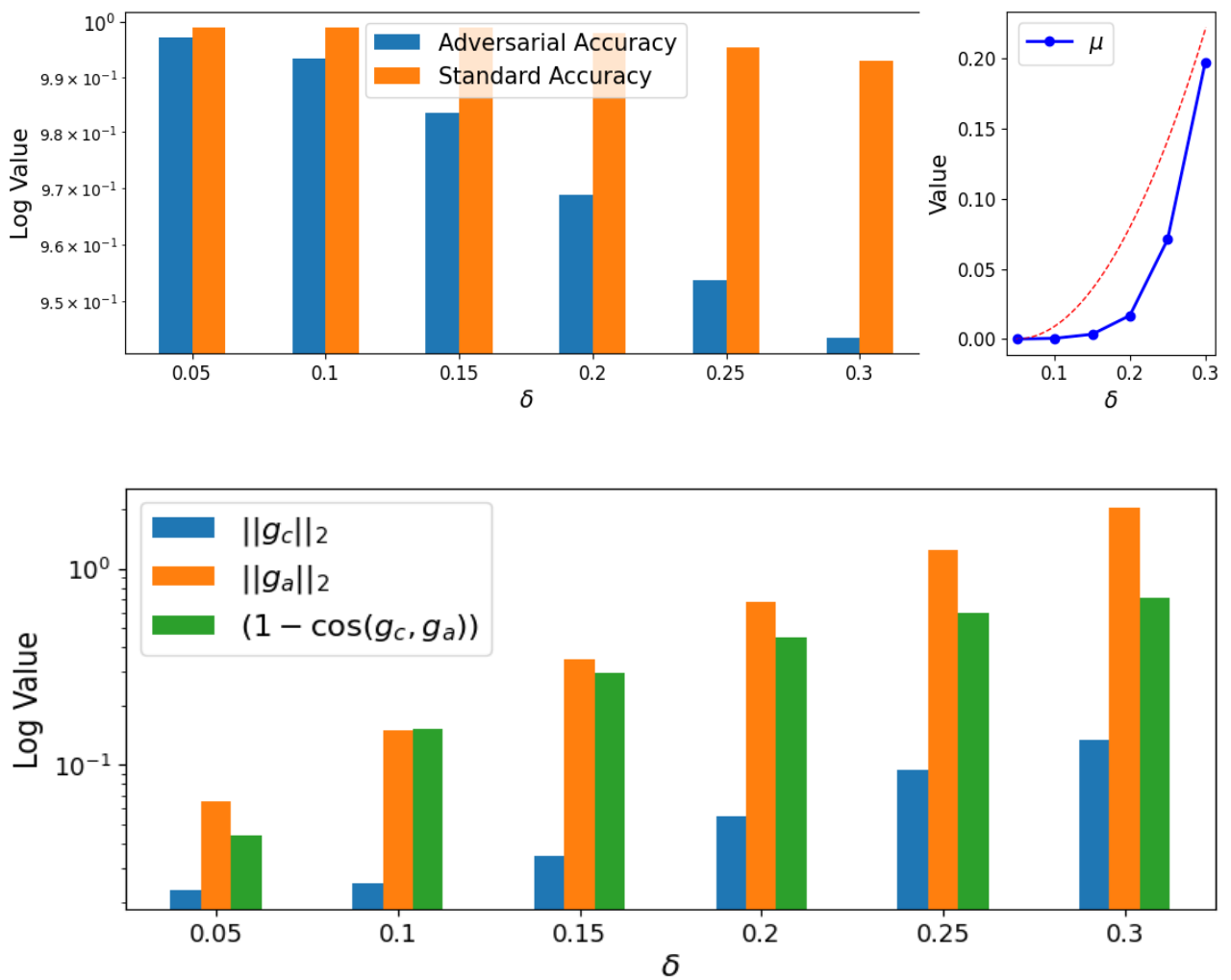}
    \caption{}
    \label{fig:Toy(b)}
  \end{subfigure}
  \caption{
  The experimental results of conducting Vanilla AT with $\lambda=0.5$ for a binary classification task on our MNIST-crafted data. In \textbf{\cref{fig:Toy(a)}}, each subfigure is the tSNE~\citep{hinton2002tsne} visualization displaying the distribution of adversarial gradients~($\ga$) and standard gradients~($\gc$) for various training samples at the final epoch with different attack budgets~($\delta = [0.05,0.1,0.15,0.2,0.25,0.3]$). In \textbf{\cref{fig:Toy(b)}}, the upper bar chart shows the standard and adversarial accuracy on testing set with different $\delta$ similar to \cref{fig:Toy(a)}. The upper left line chart shows the relation between the $\mu=||\gc||_{2} \cdot ||\ga||_{2} \cdot (1-\cos(\gc,\ga))$ and $\delta$, where the red line is the theoretical upper bound presented in Theorem 1. For decomposing $\mu$, lower bar chart shows the relation between $\delta$ and $||g_{c}||_{2}$/$||g_{a}||_{2}$/$(1-\cos(g_{a},g_{c}))$, respectively.
  }
  \label{fig:Toy}
\end{figure}
Compared to the experiments on real-world datasets, this synthetic experiment offers a distinct advantage in terms of the ability to analytically solve the inner maximization. For real-world datasets, only numerical solutions can be derived using gradient-based attacks~(e.g. PGD) during AT. These numerical solutions sometimes are not promising due to gradient masking~\citep{athalye2018obfuscated,papernot2017practical}. On the contrary, our synthetic experiments can ensure a high-quality solution for inner maximization, eliminating the potential effect of experimental results caused by some uncertainties such as gradient masking.

Under the circumstance of a simple logistic regression model with analytical solution for inner maximization, the hybrid loss for Vanilla AT can be presented as \cref{Eq: Toy_Linear}, where $\exp()$ denotes the exponential function. The details of getting the analytical solution for inner maximization will be presented in the appendix. 
\begin{align}
     \min_{\theta} \sE_{(x,y) \sim \mathcal{D}} &[\lambda\log(1+\exp(-y\cdot (w^{T}x+b) +\delta||w||_{1})) \notag  \\
     & + (1-\lambda)\log(1+\exp(-y\cdot (w^{T}x+b))] \label{Eq: Toy_Linear}
\end{align}
\cref{fig:Toy} illustrates the results of this synthetic experiment. By TSNE, \cref{fig:Toy(a)} visualizes the distributions of $\ga$ and $\gc$ for different training samples in the last training epoch. With the increase of attack budget $\delta$, these two distributions are progressively fragmented, meaning $\ga$ and $\gc$ become more different.

For \cref{fig:Toy(a)}, it is the tSNE visualization depicting the distributions of $\ga$ and $\gc$ for different training samples across varying $\delta$. Particularly, the distributions of $\ga$ and $\gc$ begin to segregate more distinctly as $\delta$ becomes larger, concomitant with the increasing gradient conflict $\mu$. Furthermore, the bar chart \cref{fig:Toy(b)} reveals a decline in both standard and adversarial accuracies with increasing $\delta$ and $\mu$. This trend indicates that the larger gradient conflict can harm the model's performances on both standard and adversarial accuracies. The subfigure on the right side of \cref{fig:Toy(b)} shows an almost quadratic growth relationship between $\mu$ and $\delta$, the red line is the theoretical upper bound derived from Theorem 1, demonstrating the effectiveness of Theorem 1 empirically.

\begin{figure*}[t]
\begin{subfigure}[b]{0.5\textwidth}
    \centering
    \includegraphics[width=\textwidth]{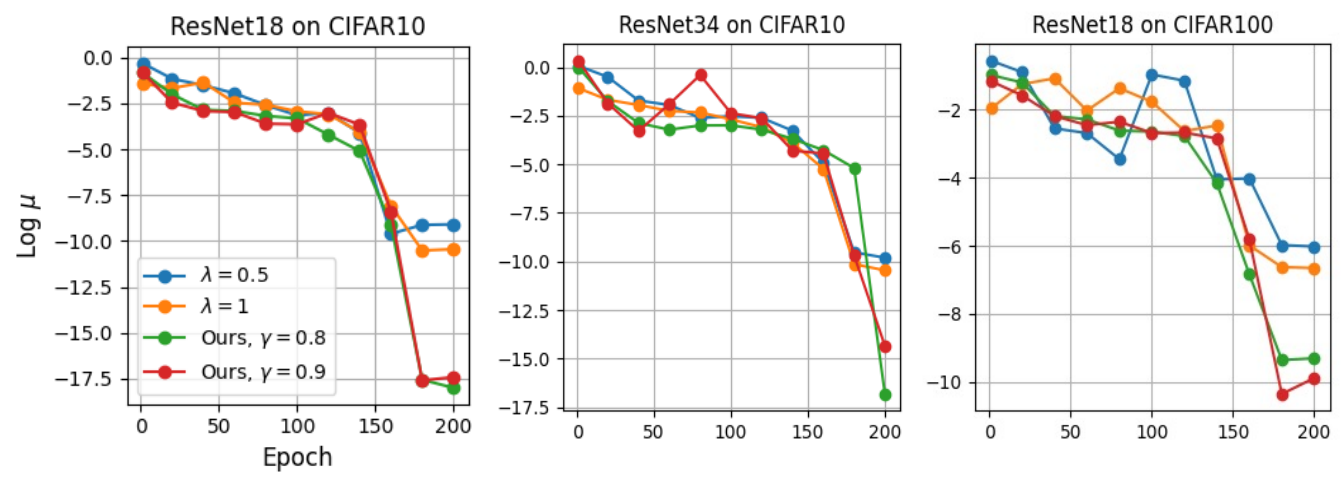}
    \caption{Different Model Architectures and Datasets}
    \label{fig:Conflict_Difdata}
  \end{subfigure}
    \begin{subfigure}[b]{0.5\textwidth}
    \centering
    \includegraphics[width=\textwidth]{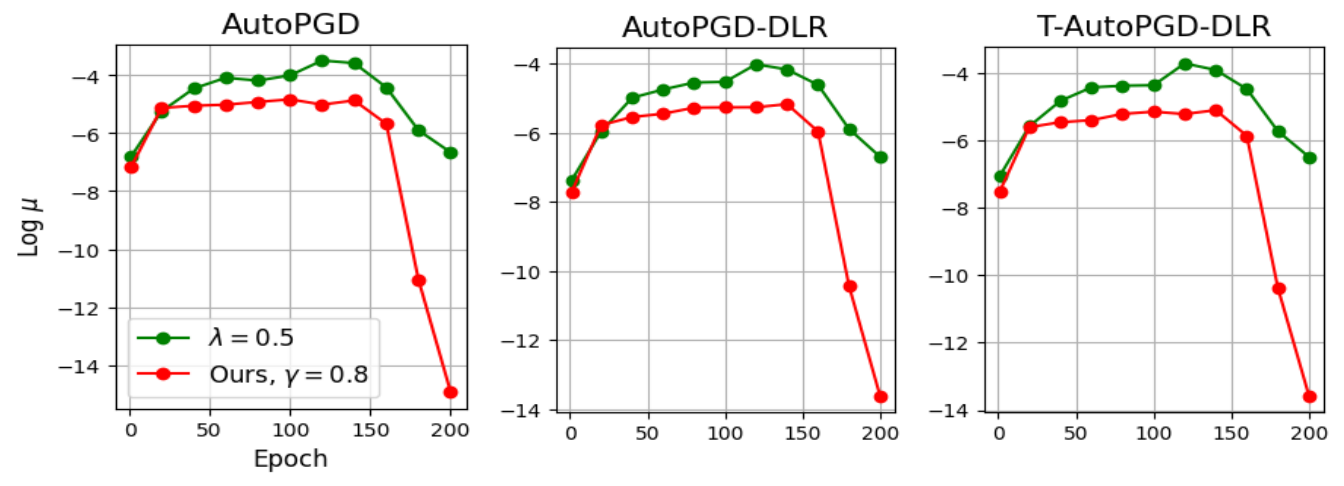}
    \caption{Different Attack Methods in AT}
    \label{fig:Conflict_Difatk}
  \end{subfigure}
    \caption{Results of gradient conflict metric $\mu$ on real-world datasets. \cref{fig:Conflict_Difdata} illustrates the results of $\mu$ among different real-world datasets~(CIFAR10/CIFAR100) and model architectures~(ResNet18/ResNet34), where the attack method used in AT is PGD. \cref{fig:Conflict_Difatk} shows the results of $\mu$ for different attack methods~(AutoPGD/AutoPGD-DLR/T-AutoPGD-DLR) during AT, conducted on CIFAR10 with ResNet18.}
    \label{fig:Conflict}
\end{figure*}

\textbf{Experiments on Real-world Datasets. }Beyond the synthetic experiment, we also conduct experiments on real-world datasets such as CIFAR10/CIFAR100, and we also observe the gradient conflict during AT. \cref{fig:Conflict} shows that such a conflict exists varying from different datasets, model architectures, and attack methods used in AT, and our method~($\gamma=0.8$, $\gamma=0.9$), which will be introduced in the next section, can consistently alleviate the conflict compared to the Vanilla AT~($\lambda=0.5$, $\lambda=1$).

%% file: Sections/Method.tex
\section{Methodology}
\label{sec:Method}

As we mentioned in \cref{sec:Con}, the trade-off between standard and adversarial accuracy is profoundly influenced by the gradient conflict $\mu$~(\cref{fig:Toy}). The vanilla AT, which employs a linear trade-off factor $\lambda$ to combine clean and adversarial loss~(as seen in \cref{Eq: AdverT_Linear}), does not adequately address the issue of gradient conflict. 

Based on this observation, we introduce Conflict-aware Adversarial Training~(CA-AT) as a new trade-off paradigm for AT. The motivation of CA-AT is that the gradient conflict in AT can be alleviated by generally conducting operations on the adversarial gradient $\ga$ and the standard gradient $\gc$ during the training process, and such an operation should guarantee the standard accuracy because its priority is higher adversarial accuracy. Inspired by existing works related to gradient operation~\cite{{yu2020gradient,yu2020gradient,liu2021conflict,chaudhry2018agem,mansilla2021domain}}, CA-AT employs a pre-defined trade-off factor $\gamma$ as the goal of cosine similarity between $\gc$ and $\ga$. In each iteration, instead of updating parameter $\theta$ by linearly weighted-averaged gradient $g_{\circ}$, CA-AT utilizes $g_{*}$ to update $\theta$ as \cref{Eq: GradietSur}
\begin{equation}
g_{*} = 
\begin{cases}
\ga + \frac{||\ga||_{2}(\gamma\sqrt{1-\phi^2} - \phi\sqrt{1-\gamma^2})}{||\gc||_{2}\sqrt{1-\gamma^2}} \gc, \quad \phi \leq \gamma \\
\gc, \quad \phi > \gamma
\end{cases} \label{Eq: GradietSur}
\end{equation}
where $\phi=\cos(\ga,\gc)$ is the cosine similarity between standard gradient $\gc$ and adversarial gradient $\ga$. The intuitive explanation of \cref{Eq: GradietSur} is depicted in \cref{fig:arch}. For each optimization iteration, if $\phi$ is less than $\gamma$, then $g_{*}$ is produced by projecting $\ga$ onto the cone of $\gc$ at an angle $\arccos{(\gamma)}$. If $\phi > \gamma$, we will use the standard gradient $\gc$ to optimize $\theta$, because we need to guarantee standard accuracy when the conflict is not quite serious. 

The mechanism behind \cref{Eq: GradietSur} is straightforward. It mitigates the gradient conflict in AT by ensuring that $\gc$ is consistently projected in a direction close to $\ga$. Considering an extreme case that $\gc$ and $\ga$ are diametrically opposite~($\ga=-\gc$), in such a scenario, if we produce the gradient by Vanilla AT as $g_{\circ} = \gc + \gc$, $g_{\circ}$ will be a zero vector and the optimization process will be stuck. On the other hand, $g_{*}$ will align closely to $\gc$ within $\gamma$, avoiding $\theta$ to be stuck in a suboptimal point. 

Furthermore, under the condition of $\phi \leq \gamma$, we find that CA-AT can also be viewed as a convex combination for standard and adversarial loss with a conflict-aware trade-off factor $\lambda^{*}$ as $\mathcal{L} = \Lc + \lambda^{*}\Ladv$, where $\lambda^{*} = \frac{||\ga||_{2}(\gamma\sqrt{1-\phi^2} - \phi\sqrt{1-\gamma^2})}{||\gc||_{2}\sqrt{1-\gamma^2}}$. Intuitively, $\lambda^{*}$ increases with the decreasing of $\phi$, which means we lay more emphasis on the standard loss when the conflict becomes more serious, and the hyperparameter $\gamma$ here serves a role of temperature to control the intensity of changing to $\lambda^{*}$. 

\begin{algorithm}
\caption{CA-AT}
\renewcommand{\algorithmicrequire}{\textbf{Input:}}
\renewcommand{\algorithmicensure}{\textbf{Output:}}
\begin{algorithmic}[1]
\Require Training dataset $D$, Loss function $\mathcal{L}$, Perturbation budget $\delta$, Training epochs $N$, Initial model parameter $\theta_{1}$, Projection margin threshold $\gamma$, learning rate $lr$
\Ensure Trained model parameter $\theta_{N+1}$
\For{$t=1$ to $N$}
    \For{each batch $B$ in $D$}
        \State $\Lc = \frac{1}{|B|}\sum_{(x,y)\in B} \mathcal{L}(x,y;\theta_{t})$ 
        \State $\Ladv = \frac{1}{|B|}\sum_{(x,y)\in B} \max_{||\epsilon||_{\infty}\leq \delta}\mathcal{L}(x+\epsilon,y;\theta_{t})$ 
        \State $\gc,\ga = \nabla_{\theta_{t}} \Lc,\nabla_{\theta_{t}} \Ladv$
        \State $\phi = \cos(\gc,\ga)$        
        \If{$\phi < \gamma$}
            \State $g_{*} = \ga + \frac{||\ga||_{2}(\gamma\sqrt{1-\phi^2} - \phi\sqrt{1-\gamma^2})}{||\gc||_{2}\sqrt{1-\gamma^2}}g_{c}$
        \Else
            \State $g_{*} = \gc$
        \EndIf
        \State $\theta_{t} = \theta_{t} - lr*g_{*}$
    \EndFor
    \State $\theta_{t+1}=\theta_{t}$
\EndFor
\end{algorithmic}
\label{alg: CA-AT}
\end{algorithm}
The pseudo-code of the CA-AT is shown as \cref{alg: CA-AT}. In each training batch $B$, we calculate both standard loss $\Lc$ and adversarial loss $\Ladv$. By evaluating and adjusting the alignment
between standard gradient $\gc$ and adversarial gradient $\ga$, the algorithm ensures the model not only performs well via standard samples but also maintains robustness against designed perturbations. This adjustment is made by modifying the adversarial gradient $\ga$ to better align with the standard gradient $\gc$ based on the projection margin threshold $\gamma$, where $g_{*}$ is produced to optimize the model parameter $\theta_{t}$ in each round $t$.

%% file: Sections/exp.tex
\section{Experimental Results \& Analysis}
\label{sec: Exp}

\begin{figure*}[t]
   \begin{subfigure}[b]{0.5\textwidth}
    \centering
    \includegraphics[width=\textwidth,height=0.45\textwidth]{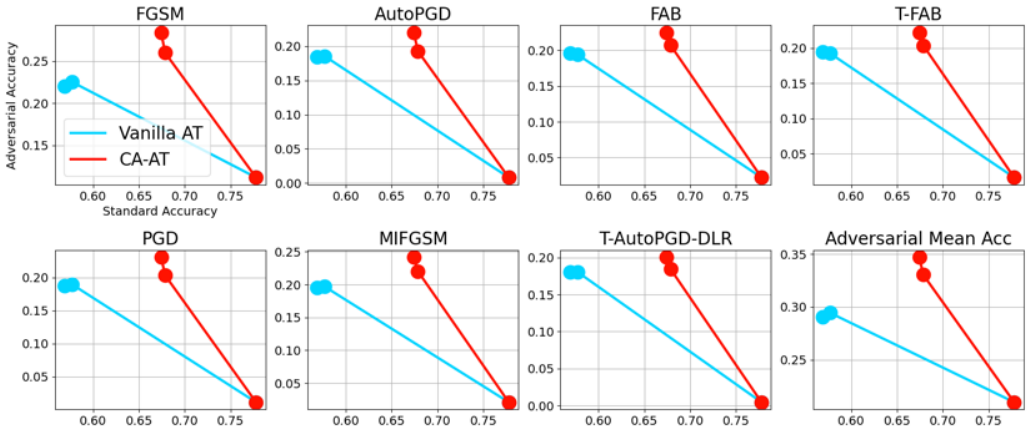}
    \caption{CUB-Bird}
    \label{fig:Adapter_CUB}
  \end{subfigure}
    \begin{subfigure}[b]{0.5\textwidth}
    \centering
    \includegraphics[width=\textwidth,height=0.45\textwidth]{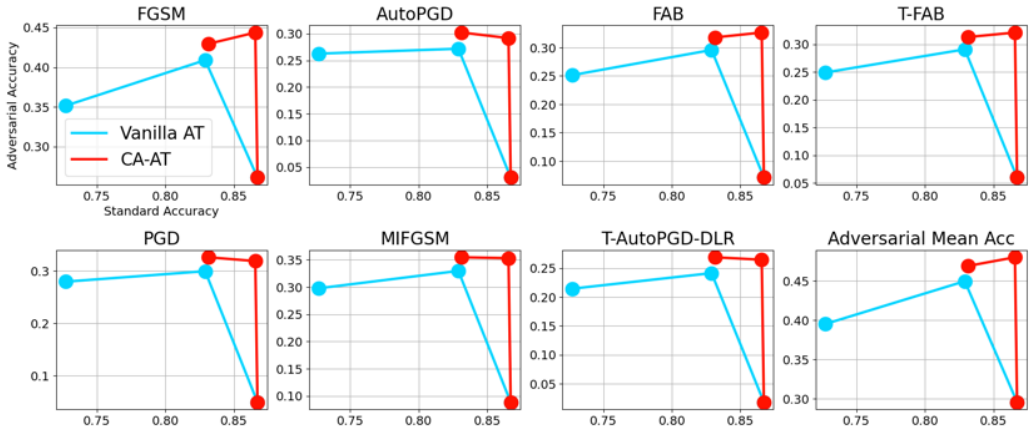}
    \caption{Stanford Dogs}
    \label{fig:Adapter_Dogs}
  \end{subfigure}
  \begin{subfigure}[b]{0.5\textwidth}
    \centering
    \includegraphics[width=\textwidth,height=0.45\textwidth]{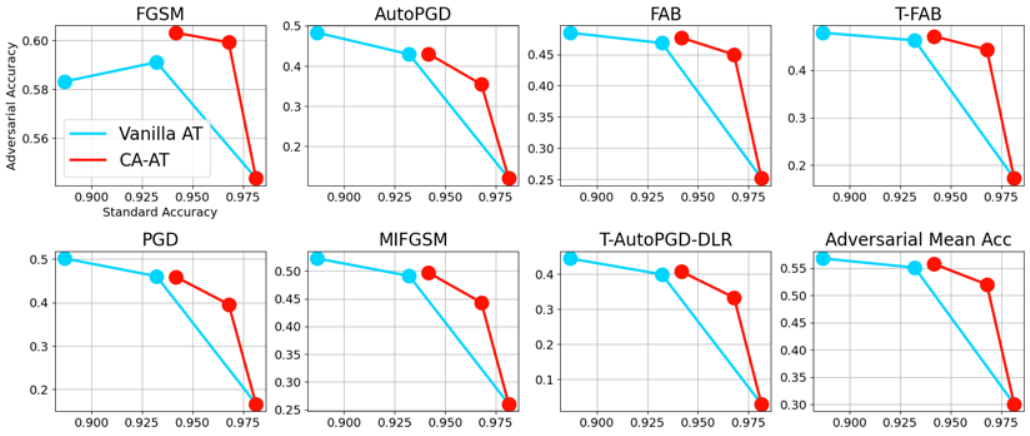}
    \caption{CIFAR10}
    \label{fig:Adapter_CIFAR10}
  \end{subfigure}
    \begin{subfigure}[b]{0.5\textwidth}
    \centering
    \includegraphics[width=\textwidth,height=0.45\textwidth]{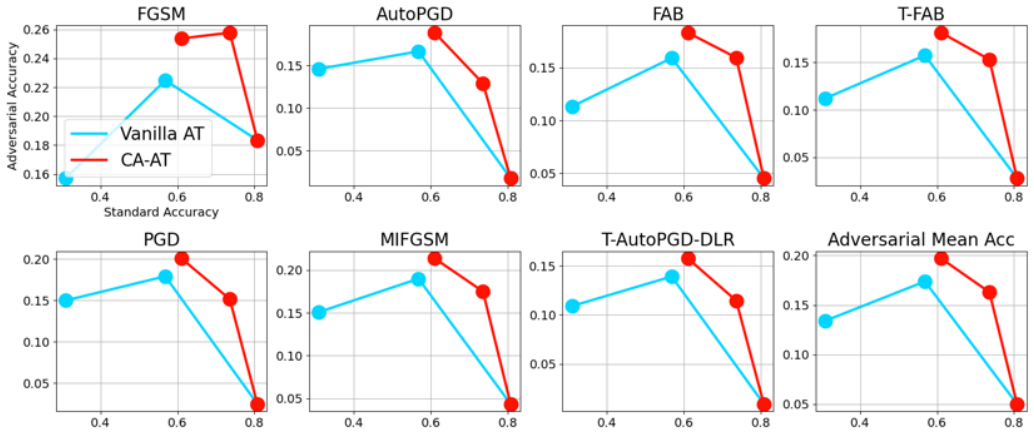}
    \caption{CIFAR100}
    \label{fig:Adapter_CIFAR100}
  \end{subfigure}
    \caption{SA-AA Fronts for Adversarial PEFT on Swin-T using Adapter.}
    \label{fig:Adapter}
\end{figure*}

\begin{figure}
    \centering
    \includegraphics[width=\textwidth]{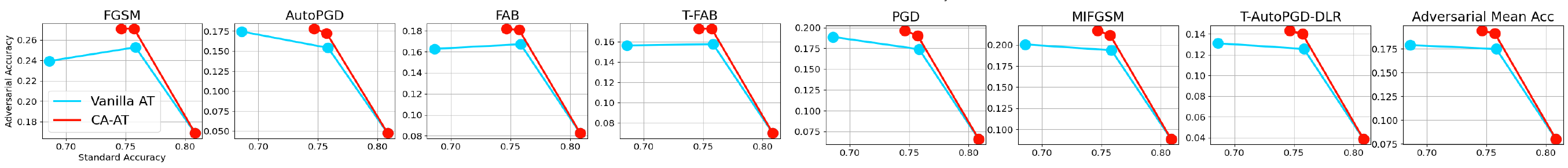}
    \caption{SA-AA Fronts for Adversarial PEFT on ViT using Adapter on Stanford Dogs.}
    \label{fig:Adapter_ViT}
\end{figure}


\begin{figure*}[t!]
    \centering
    \begin{minipage}{\textwidth}
        \centering
        \includegraphics[width=\textwidth]{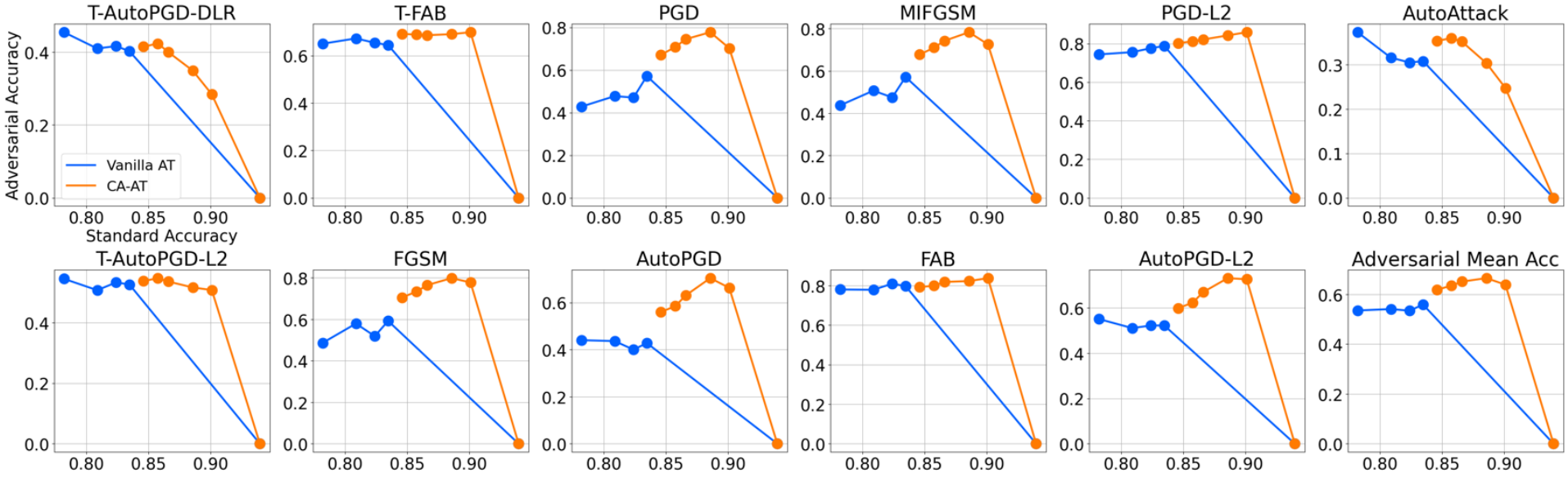}
        \subcaption{Cross Entropy}
         \label{fig:cifar10_adver_res18}
    \end{minipage}%
    \hfill
    \centering
    \begin{minipage}{\textwidth}
        \centering
        \includegraphics[width=\textwidth]{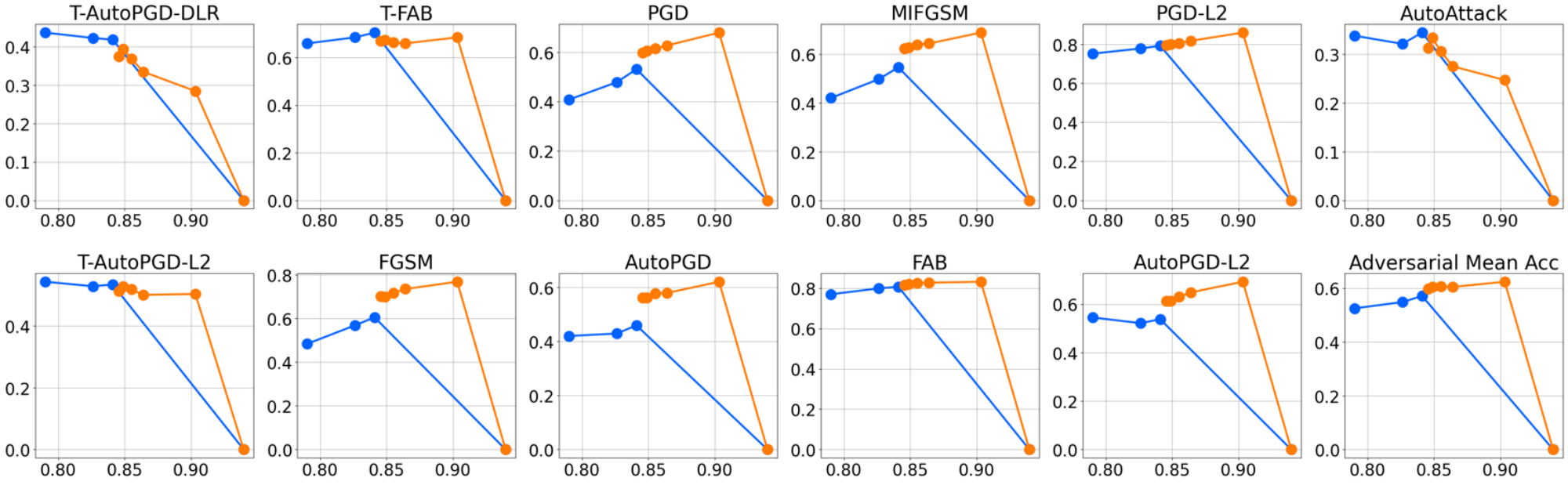}
        \subcaption{TRADES}
        \label{fig:cifar10_trades_res18}
    \end{minipage}%
    \hfill
    \centering
    \begin{minipage}{\textwidth}
        \centering
        \includegraphics[width=\textwidth]{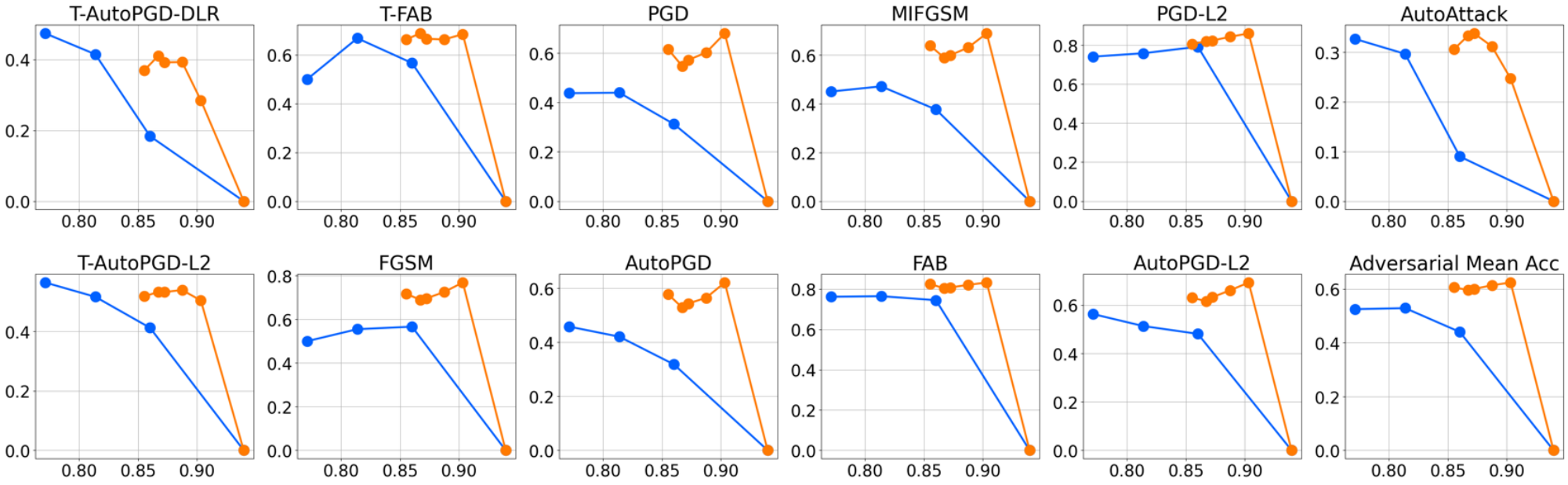}
        \subcaption{CLP}
        \label{fig:cifar10_clp_res18}
    \end{minipage}%
\label{fig:cifar10_res18}
\caption{SA-AA Fronts for Adversarial Training from Scratch on CIFAR10 using ResNet18 with Different Adversarial Loss Functions including Cross Entropy, TRADES, and CLP.}
\end{figure*}

In this section, we demonstrate the effectiveness of CA-AT for achieving better trade-off results compared to Vanilla  AT. We conduct experiments on adversarial training from scratch and adversarial PEFT among various datasets and model architectures. Besides, motivated by Theorem 1, we evaluate CA-AT by involving adversarial samples with a larger budget in training. Experimental results show that CA-AT can boost the model's robustness by handling adversarial samples with a larger budget, while Vanilla  AT fails.
\subsection{Experimental Setup}
\textbf{Datasets and Models. }We evaluate our proposed method on various image classification datasets including CIFIAR10~\citep{krizhevsky2009learning}, CIFIAR100~\citep{krizhevsky2009learning}, CUB-Bird~\citep{cubbird}, and StanfordDogs~\citep{dogs}. The model architectures we utilized to train from scratch on CIFAR10 and CIFAR100 are ResNet18, ResNet34~\citep{he2016deep}, and WideResNet28-10~(WRN-28-10)~\citep{zagoruyko2016wide}. We set the value of running mean and running variance in each Batch Normalization block into false as a trick to boost adversarial robustness~\citep{wang2022removing,walter2022fragile}. For experiments on PEFT, we fine-tune Swin Transformer~(Swin-T)~\citep{liu2021swin} and Vision Transformer~(ViT)~\citep{dosovitskiy2020vit} by using Adapter~\citep{pfeiffer2020adapter1,pfeiffer2021adapter2}, which fine-tunes the large pretrained model by inserting a small trainable module into each block. Such a module adapts the internal representations for specific tasks without altering the majority of the pretrained model’s parameters. Both Swin-T and ViT are pretrained adversarially~\citep{dong2020adverpretrain} on ImageNet. For the experiments on ResNet, we set the resolution of input data as $32\times 32$, and use resolution as $224\times 224$ for the PEFT experiments on Swin-T and ViT.

\textbf{Hyper-parameters for AT. }For adversarial training from scratch, we use the PGD attack with $\delta=8/255$ with step size $2/255$ and step number $10$. For the optimizer, we use SGD with momentum as $0.9$ and the initial learning rate as $0.4$. We use the one-cycle learning rate policy~\citep{smith2019super} as the dynamic adjustment method for the learning rate within 200 epochs. The details of hyperparameter setup for adversarial PEFT will be shown in Appendix. Generally, we use a sequence of operations as random crop, random horizontal flip, and random rotation for data augmentation. For a fair comparison, we maintain the same hyper-parameters across experiments for vanilla AT and CA-AT on both adversarial training from scratch and PEFT.

\textbf{Evaluation. }We evaluate adversarial robustness by reporting the accuracies against extensive adversarial attacks constrained by $L_{\infty}$ and $L_{2}$. For attacks bounded by $L_{\infty}$ norm, we selected most representative methods including PGD~\citep{madry2018towards}, AutoPGD~\citep{croce2020Auto}, FGSM~\citep{goodfellow2014explaining}, MIFGSM~\citep{dong2018boosting}, FAB~\citep{croce2020fab}, and AutoAttack~\citep{croce2020Auto}. Besides, we also conducted the targeted adversarial attacks, where they are denoted as a 'T-' as the prefix~(e.g. T-AutoPGD). For all the targeted adversarial attacks, we set the number of classes as $10$. Attacks bounded by $L_{2}$ norm are denoted as '-L2' in suffix~(e.g. AutoPGD-L2). Besides, we apply attacks with different loss functions such as cross entropy~(AutoPGD) and difference of logits ratio~(AutoPGD-DLR), to avoid the `fake' adversarial examples caused by gradient vanishing~\citep{athalye2018obfuscated}. 

To measure the quality of trade-off between standard accuracy~(SA) and adversarial accuracy~(AA), we define \textbf{SA-AA front} as an empirical Pareto front for SA and AA. We draw this front by conducting different $\lambda$ on Vanilla  AT and different $\gamma$ on CA-AT. 
\subsection{Experimental Results on PEFT}

\textbf{CA-AT offers the better trade-off on adversarial PEFT. }\cref{fig:Adapter} shows the SA-AA fronts on fine-tuning robust pretrained Swin Transformer on CUB-Bird and StanfordDogs by using Adapter. We set $\lambda=[0,0.5,1]$ for Vanilla  AT and $\gamma=[0.8,0.9,1]$ for CA-AT. The red data points for CA-AT are positioned in the upper right area relative to the blue points for Vanilla AT. It shows that CA-AT can consistently attain better standard and adversarial accuracy compared to the Vanilla AT across different datasets. Besides, we observed that on fine-grained datasets such as CUB-Bird and Stanford Dogs, the superiority of CA-AT is more significant compared to the results on normal datasets.

\textbf{Results for CA-AT with Different Pretrained Models. }\cref{fig:Adapter_ViT} shows that CA-AT can also boost the trade-off performance on ViT. The main difference between these two models is that, ViT treats image patches as tokens and processes them with a standard transformer architecture~\cite{vaswani2017attention}, while Swin-T uses shifted windows for hierarchical feature merging. While ViT applies global attention directly on image patches, Swin Transformer applies local attention within windows and uses a hierarchical approach to better handle larger and more detailed images. The superiority of CA-AT on ViT is not as significant as it is on Swin-T~(\cref{fig:Adapter_Dogs}), but it still can gain better standard and adversarial accuracy compared to Vanilla AT.
 
\subsection{Experimental Results on Training from Scratch}
\begin{figure*}[t]
   \begin{subfigure}[b]{0.5\textwidth}
    \centering
    \includegraphics[width=\textwidth]{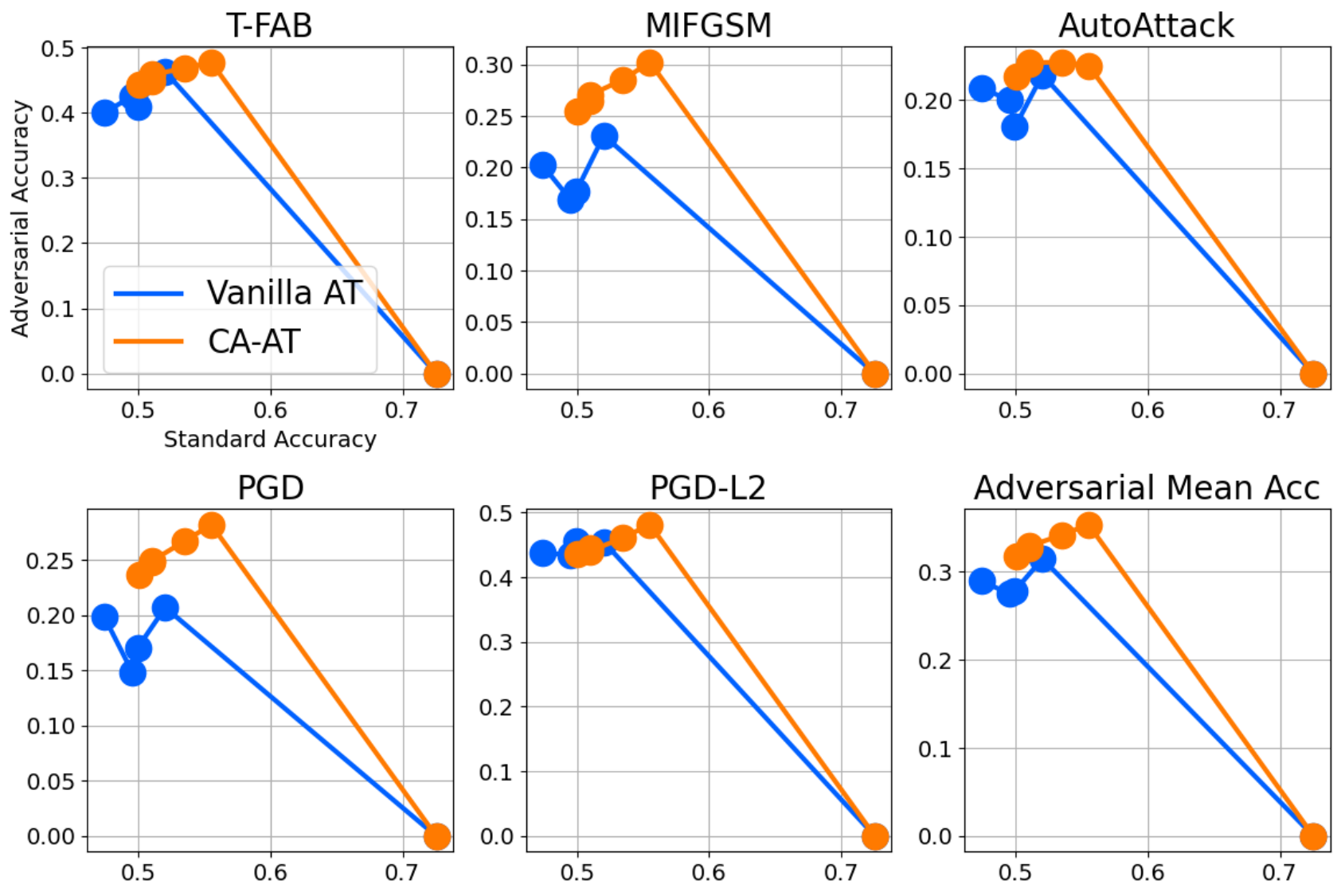}
    \caption{ResNet18}
    \label{fig:cifar100_res18}
  \end{subfigure}
    \begin{subfigure}[b]{0.5\textwidth}
    \centering
    \includegraphics[width=\textwidth]{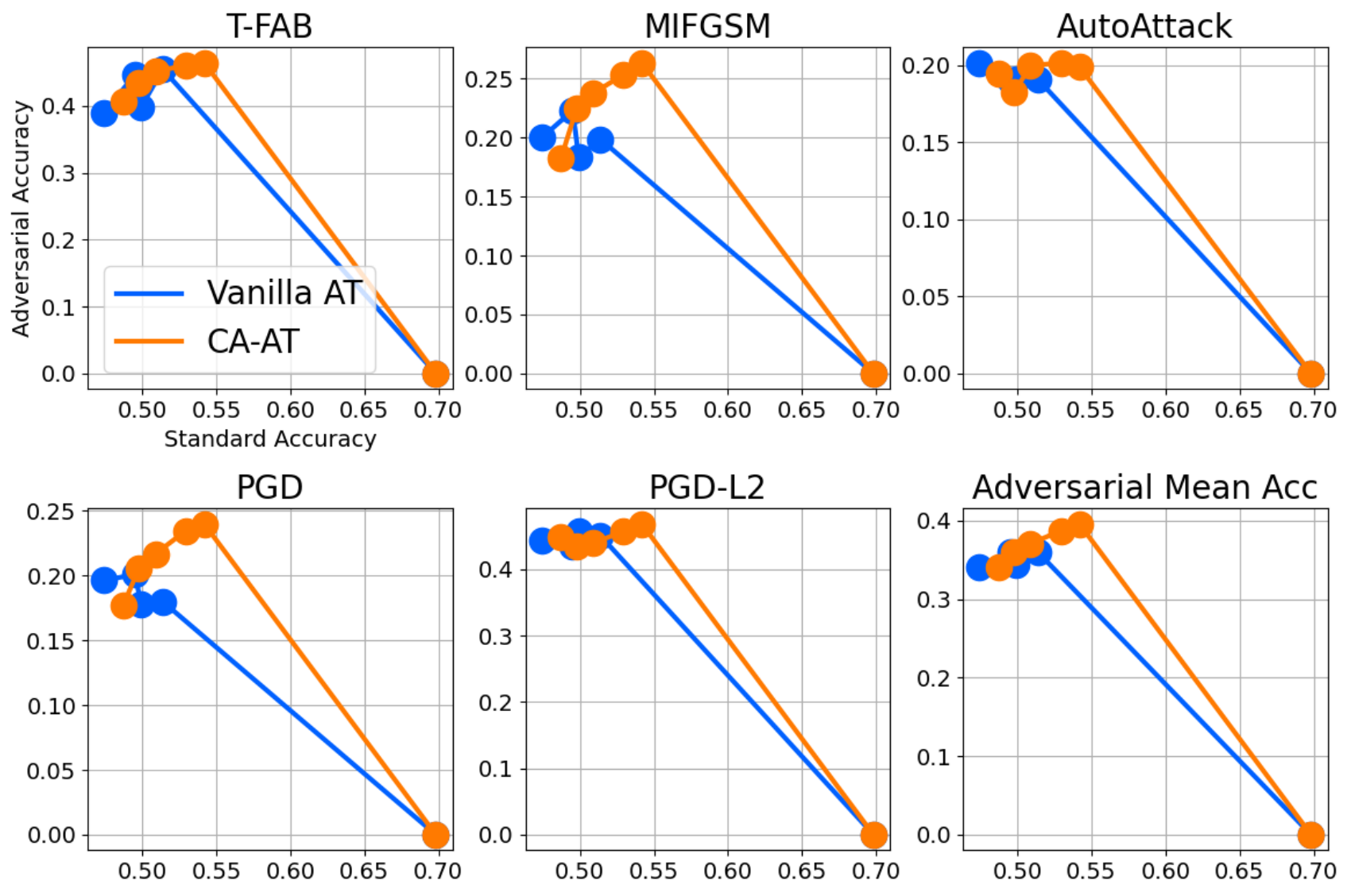}
    \caption{ResNet34}
    \label{fig:cifar100_res34}
  \end{subfigure}
  \caption{SA-AA Fronts for Adversarial Training from Scratch on CIFAR100.}
\label{fig:cifar100}
\end{figure*}


\begin{table*}[t]
\centering
\resizebox{1\textwidth}{0.102\textwidth}{%
\begin{tabular}{c|c|cc|cc|cc|cc|cc|cc|cc}
\hline
\multicolumn{1}{l|}{}     & \multicolumn{1}{l|}{} & \multicolumn{2}{c|}{Standard Accuracy}                     & \multicolumn{2}{c|}{PGD}        & \multicolumn{2}{c|}{AutoPGD}    & \multicolumn{2}{c|}{MIFGSM}     & \multicolumn{2}{c|}{FAB}          & \multicolumn{2}{c|}{T-FAB}      & \multicolumn{2}{c}{FGSM}        \\ \hline
\multicolumn{1}{l|}{}     & $p=\infty$            & CA-AT                     & Vanilla  AT           & CA-AT    & Vanilla  AT & CA-AT    & Vanilla  AT & CA-AT    & Vanilla  AT & CA-AT    & Vanilla  AT   & CA-AT    & Vanilla  AT & CA-AT    & Vanilla  AT \\ \hline
\multirow{4}{*}{ResNet18} & 8/255                 & \multirow{4}{*}{\textbf{0.8659}} & \multirow{4}{*}{0.8239} & \textbf{0.7442} & 0.4703        & \textbf{0.6301} & 0.3996        & \textbf{0.7419} & 0.4745        & \textbf{0.8177} & 0.809           & \textbf{0.6861} & 0.6538        & \textbf{0.7649} & 0.519         \\
                          & 16/255                &                                  &                         & \textbf{0.7311} & 0.4248        & \textbf{0.5555} & 0.2486        & \textbf{0.7233} & 0.4225        & 0.7475          & \textbf{0.78}   & \textbf{0.5445} & 0.5104        & \textbf{0.7435} & 0.4387        \\
                          & 24/255                &                                  &                         & \textbf{0.7189} & 0.405         & \textbf{0.4886} & 0.1963        & \textbf{0.7182} & 0.413         & 0.6858          & \textbf{0.7333} & \textbf{0.4599} & 0.4783        & \textbf{0.7235} & 0.403         \\
                          & 32/255                &                                  &                         & \textbf{0.7033} & 0.3877        & \textbf{0.4455} & 0.1589        & \textbf{0.7182} & 0.413         & 0.6402          & \textbf{0.6836} & \textbf{0.4044} & 0.4507        & \textbf{0.7066} & 0.379         \\ \hline
\multicolumn{1}{l|}{}     &                       & \multicolumn{2}{c|}{Standard Accuracy}                     & \multicolumn{2}{c|}{PGD}        & \multicolumn{2}{c|}{AutoPGD}    & \multicolumn{2}{c|}{MIFGSM}     & \multicolumn{2}{c|}{FAB}          & \multicolumn{2}{c|}{T-FAB}      & \multicolumn{2}{c}{FGSM}        \\ \hline
                          & $p=\infty$            & CA-AT                     & Vanilla  AT           & CA-AT    & Vanilla  AT & CA-AT    & Vanilla  AT & CA-AT    & Vanilla  AT & CA-AT    & Vanilla  AT   & CA-AT    & Vanilla  AT & CA-AT    & Vanilla  AT \\ \hline
\multirow{4}{*}{ResNet34} & 8/255                 & \multirow{4}{*}{\textbf{0.8753}} & \multirow{4}{*}{0.8305} & \textbf{0.8098} & 0.5973        & \textbf{0.7285} & 0.4417        & \textbf{0.8111} & 0.5983        & \textbf{0.8247} & 0.8068          & \textbf{0.7274} & 0.6951        & \textbf{0.8149} & 0.5327        \\
                          & 16/255                &                                  &                         & \textbf{0.8034} & 0.5756        & \textbf{0.6793} & 0.3395        & \textbf{0.8077} & 0.5791        & \textbf{0.7738} & 0.7613          & \textbf{0.6013} & 0.5762        & \textbf{0.7916} & 0.2762        \\
                          & 24/255                &                                  &                         & \textbf{0.7957} & 0.5602        & \textbf{0.6445} & 0.2859        & \textbf{0.8067} & 0.5743        & \textbf{0.7307} & 0.6937          & \textbf{0.5142} & 0.5174        & \textbf{0.7743} & 0.1428        \\
                          & 32/255                &                                  &                         & \textbf{0.785}  & 0.5443        & \textbf{0.6165} & 0.2498        & \textbf{0.8067} & 0.5743        & \textbf{0.6918} & 0.6221          & \textbf{0.4424} & 0.4822        & \textbf{0.7616} & 0.088         \\ \hline
\end{tabular}%
}
\caption{Evaluation results on CIFAR10 for CA-AT~($\gamma=0.8$) and Vanilla  AT~($\lambda=0.5$) across different $L_{\infty}$-based attacks with various values of budget $\delta$.}
\label{tab:dif_bounds_linf}
\end{table*}

\begin{figure}[t]
    \centering
    \includegraphics[width=\textwidth]{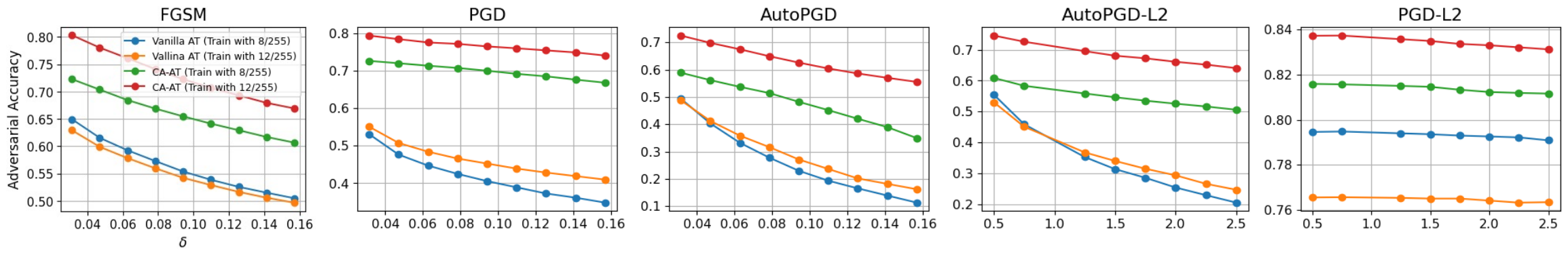}
    \caption{Results for Vanilla  AT and CA-AT trained on adversarial samples with two different budget values~($\delta=8/255$,$\delta=12/255$) on CIFAR10 with ResNet18. We evaluate the adversarial accuracy among different adversarial attacks with different budget values $\delta$ denoting as the x-axis.}
    \label{fig:DifTrainBounds}
\end{figure}
\textbf{CA-AT results in better trade-off with different adversarial loss functions. }
\cref{fig:cifar10_adver_res18} visualizes SA-AA fonts from experiments using vanilla AT with $\lambda=[0,0.25,0.5,0.75,1]$ and CA-AT with $\gamma=[0.7,0.75,0.8,0.85,0.9,1]$ on CIFAR10. In this figure, most orange data points~(CA-AT) lie in the upper right space of blue points~(Vanilla AT), indicating that CA-AT offers a better empirical Pareto front for the trade-off between standard accuracy and adversarial accuracy. Moreover, \cref{fig:cifar10_clp_res18} and \cref{fig:cifar10_trades_res18} show CA-AT can also consistently boost the adversarial accuracy for different adversarial loss functions used in AT such as TRADES~\citep{zhang2019theoretically} and CLP~\citep{kannan2018logitpair}. For the experiments on CIFAR100, we selected the strongest and most representative attack methods to evaluate the model’s robustness, including targeted attack~(T-FAB), untargeted attacks~(PGD, MIFGSM), $L_{2}$-norm attack~(T-PGD), and ensemble attack~(AutoAttack). Showing the trade-off results on CIFAR100 in \cref{fig:cifar100_res18}~(ResNet18) and \cref{fig:cifar100_res34}~(ResNet34), the performance gain of CA-AT is more limited compared to the one on CIFAR10, but it can still achieve better performance on standard accuracy and adversarial accuracy against various adversarial attacks. 

\textbf{CA-AT is more robust to adversarial attacks with larger budget values.}
We evaluate adversarial precision through various adversarial attacks with different attack budget values $\delta$, to demonstrate the superiority of our model over Vanilla AT under various intensities of adversarial attacks. We applied both Vanilla AT and CA-AT to ResNet18 on CIFAR10, and the results about $L_{\infty}$-based attacks are shown in \cref{tab:dif_bounds_linf}. In \cref{tab:dif_bounds_linf}, although our CA-AT achieves slightly lower adversarial accuracy against FAB when $\delta$ is larger than $8/255$, it outperforms the Vanilla  AT in both standard accuracy and adversarial accuracy on any other attack methods~(e.g. AutoPGD, MIFGSM, and T-FAB) with different budget $\delta$. It clearly illustrates that, compared to Vanilla AT, CA-AT can enhance the model's adversarial robustness
ability to resist stronger adversarial attacks with larger budget $\delta$.

\textbf{CA-AT enables AT via stronger adversarial examples. }In our toy experiment~(\cref{fig:Toy}) and Theorem 1, the conflict $\mu$ would be more serious if we utilize adversarial examples with larger attack budget $\delta$ during AT. It implies that Vanilla AT cannot handle stronger adversarial examples during training because of the gradient conflict. In \cref{fig:DifTrainBounds}, we visualize the results of training ResNet34 on CIFAR10 with adversarial samples produced by the same attack method~(PGD), but different attack budgets~($\delta=8/255$ and $\delta=16/255$), and evaluate the adversarial accuracies against various adversarial methods~(e.g. FGSM and PGD) with different budgets~(x-axis). Compared to the blue and orange curves~(Vanilla AT with $\delta=8/255$), it shows that Vanilla AT fails when training with the adversarial attack with a higher perturbation bound, causing a decrease in both standard and adversarial accuracy. On the contrary, CA-AT, shown as the green and red curves, can improve both standard and adversarial accuracy by involving stronger adversarial samples with larger attack budgets. 

\textbf{Experimental Results in Appendix. }More experimental results for CA-AT regarding different model architectures~(WRN-28-10), different attack methods utilized for producing adversarial samples during AT, various $L_{2}$-based attacks with different budgets, and black-box attacks can be found in \cref{sec:app:exp}. In addition, the detailed proof for Theorem 1 is included in \cref{sec:app:proof}.

%% file: Sections/appendix.tex
\section{Proof of Theorem 1}
\label{sec:app:proof}
Considering $||\epsilon||_{2}$ or $||\epsilon||_{\infty}$ is usually very small for adversarial examples, we utilize Taylor Expansion for $x$ as the approximation for the adversarial loss $\mathcal{L}(x+\epsilon;\theta)$, such that:
\begin{gather}
    \mathcal{L}(x+\epsilon;\theta) = \mathcal{L}(x;\theta) + (\frac{\partial\mathcal{L}(x;\theta)}{\partial x})^{T}\epsilon + \mathcal{O}(\|\epsilon\|^2)
\end{gather}
To derive an upper bound on the gradient conflict in the regime that $\|\epsilon\|$ gets small, we will only consider the first-order term above. We then take the derivative of both sides of the equation with respect to $\theta$ to obtain:
\begin{align}
    \ga = \gc + \frac{\partial^{2}\mathcal{L}(x;\theta)}{\partial x \partial\theta}\epsilon  = \gc + \frac{\partial \gc}{\partial x} \epsilon = \gc + H\epsilon \label{eq:app:0}
\end{align}
where $H=\frac{\partial \gc}{\partial x}\in \mathbb{R}^{d_{\theta}\times d_{x}}$. $d_{\theta}/d_{x}$ denotes the dimension of parameter $\theta$ and input data $x$. By multiplying $\ga^{T}$ and $\gc^{T}$ on the two sides of \cref{eq:app:0}, respectively, we can obtain \cref{eq:app:1} and \cref{eq:app:2} as follows.
\begin{align}
    \gc^{T}\ga &= ||\gc||_{2}^{2} + \gc^{T}H\epsilon \label{eq:app:1}\\
    ||\ga||_{2}^{2} &= \ga^{T}\gc + \ga^{T}H\epsilon \label{eq:app:2}
\end{align}
\cref{eq:app:1} minus \cref{eq:app:2}:
\begin{align}
     \gc^{T}\ga &=  \dfrac{||\ga||_{2}^{2} + ||\gc||_{2}^{2} + \epsilon^{T}H^{T}(\gc - \ga)}{2}
\end{align}
Based on \cref{eq:app:0}, we can replace $(\gc - \ga)$ as $H\epsilon$:
\begin{align}
     \gc^{T}\ga  = \dfrac{||\ga||_{2}^{2} + ||\gc||_{2}^{2} - \epsilon^{T}H^{T}H \epsilon}{2} \label{eq:app:3}
\end{align}
Recall the definition of $\mu$ as $\mu = ||\gc||_{2} \cdot ||\ga||_{2} \cdot (1-\cos(\gc,\ga))$
\begin{align}
    \mu &= ||\gc||_{2} \cdot ||\ga||_{2} \cdot (1-\cos(\gc,\ga)) \notag \\
    &= ||\gc||_{2} \cdot ||\ga||_{2} - \gc^{T}\ga \notag \quad \\
    & = \dfrac{2||\gc||_{2} \cdot ||\ga||_{2} - ||\ga||_{2}^{2} - ||\gc||_{2}^{2} + \epsilon^{T} H^{T}H\epsilon }{2} \notag \quad \text{(Use \cref{eq:app:3})} \\
    &= \frac{\epsilon^{T}\mathcal{K}(\theta,x)\epsilon -(||\gc||_{2}-||\ga||_{2})^2}{2}\leq  \frac{\epsilon^{T}\mathcal{K}(\theta,x)\epsilon}{2} \leq \frac{\lambda_{max}\epsilon^{T}\epsilon}{2}
\end{align}
where $\mathcal{K}(\theta,x) = H^{T}H$ is a symmetric and positive semi-definite matrix, and $\lambda_{max}$ is the largest eigenvalue of $K$, where $\lambda_{max}\geq 0$.

Considering two widely-used restrictions for perturbation $\epsilon$ applied in adversarial examples as $l_{2}$ and $l_{\infty}$ norm, we have:
\begin{itemize}
    \item For $||\epsilon||_{2} \leq \delta$, where $\mu\leq\dfrac{1}{2}\lambda_{max}\delta^{2}$. The upper bound of $\mu$ is $\mathcal{O}(\delta^{2})$.
    \item For $||\epsilon||_{\infty} \leq \delta$, it implies that the absolute value of each element of $\epsilon$ is bounded by $\delta$, where $\epsilon^{T}\epsilon=\sum^{d}_{i=0}\epsilon_{i}^{2}\leq d^{2}\delta^{2}$. The upper bound of $\mu$ is $\mathcal{O}(d^{2}\delta^{2})$.
\end{itemize}

\section{Analytical Solution for the Inner Maximization}
We introduce the details about how to get the analytical inner-max solution~(\cref{Eq: Toy_Linear}) for our synthetic experiment presented in \cref{sec:Con}. As we introduced in \cref{sec:Con}, consider a linear model as $f(x) = w^T x + b$ under a binary classification task where $y \in \{+1, -1\}$. The predicted probability of sample $x$ with respect to its ground truth $y$ can be defined as:
\begin{align}
p(y|x) = \frac{1}{1 + \exp(-y \cdot f(x))}
\end{align}
Then, the BCE loss function for sample $x$ can be formulated as:
\begin{align}
\mathcal{L}(f(x), y) = -\log(p(y|x)) = \log(1 + \exp(-y \cdot f(x)))
\end{align}
Consider the perturbation $\epsilon$ under the restriction of $L_{\infty}$ norm, the adversarial attack for such a linear model can be formulated as an inner maximization problem as \cref{eq:app:linear}.
\begin{align}
\max_{\|\epsilon\|_\infty \leq \delta} \log(1 + \exp(-y \cdot f(x+\epsilon))) \equiv \min_{\|\epsilon\|_\infty \leq \delta} y \cdot w^T \epsilon \quad  \label{eq:app:linear}
\end{align}
Consider the case that $y = +1$, where the $L_\infty$ norm says that each element in $\epsilon$ must have magnitude less than or equal $\delta$, we clearly minimize this quantity when we set $\epsilon_i = -\delta$ for $w_i \geq 0$ and $\epsilon_i = \delta$ for $w_i < 0$. For $y = -1$, we would just flip these quantities. That is, the optimal solution $\epsilon^*$ to the above optimization problem for the $L_\infty$ norm is expressed as \cref{eq:app:solution}.
\begin{align}
\epsilon^* = -y \cdot \delta \odot \text{sign}(w) \label{eq:app:solution}
\end{align}
where $\odot$ is the element-wise multiplication. Based on \cref{eq:app:solution}, we can formulate the adversarial loss as follows, which is as same as the adversarial loss presented in \cref{Eq: Toy_Linear}.
\begin{align}
    \mathcal{L}(f(x + \epsilon^*), y) & = \log(1 + \exp(-y \cdot w^T x - y \cdot b - y \cdot w^T \epsilon^*)) \notag \\
    &= \log(1 + \exp(-y \cdot f(x) + \delta\|w\|_1))
\end{align}

\section{Additional Experimental Results}
\label{sec:app:exp}
\begin{table*}[t]
\centering
\resizebox{\textwidth}{0.0397\textwidth}{%
\begin{tabular}{c|c|ccccccccccc|c}
\hline
                                       & Standard        & T-AutoPGD-DLR   & T-AutoPGD-L2    & T-FAB           & FGSM            & PGD             & AutoPGD         & MIFGSM          & FAB             & PGD-L2          & AutoPGD-L2      & AutoAttack      & Adversarial Mean Acc \\ \hline
$\gamma=0.8$, PGD     & \textbf{0.8659} & 0.4004          & 0.5356          & 0.6861          & \textbf{0.7649} & \textbf{0.7442} & 0.6301          & \textbf{0.7419} & \textbf{0.8177} & \textbf{0.8211} & 0.67            & \textbf{0.3517} & \textbf{0.6512}      \\
$\gamma=0.8$, PGD-DLR & 0.8646          & \textbf{0.4147} & \textbf{0.539}  & \textbf{0.7168} & 0.7452          & 0.672           & \textbf{0.6429} & 0.6864          & 0.8001          & 0.8196          & \textbf{0.6919} & 0.3260          & 0.6413               \\ \hline
$\gamma=0.9$, PGD     & \textbf{0.9009} & 0.2844          & 0.5075          & \textbf{0.6986} & 0.7781          & 0.7021          & \textbf{0.6624} & 0.7251          & \textbf{0.8371} & \textbf{0.8588} & \textbf{0.7267} & 0.2472          & 0.6389               \\
$\gamma=0.9$, PGD-DLR & 0.8923          & \textbf{0.3794} & \textbf{0.5353} & 0.6874          & \textbf{0.779}  & \textbf{0.7207} & 0.6428          & \textbf{0.7315} & 0.8229          & 0.8488          & 0.7038          & \textbf{0.2992} & \textbf{0.6501}      \\ \hline
\end{tabular}%
}
\caption{Evaluation results for CA-AT for using different inner maximization solver~(PGD/PGD-DLR) during the process of AT.}
\label{tab:dif_tr_attck}
\end{table*}

\textbf{Experimental setup on adversarial PEFT.} For the experiments on adversarial PEFT, we leverage the adversarially pretrained Swin-T and ViT downloaded from ARES\footnote{\url{https://github.com/thu-ml/ares}}. For adapter, we implement it as \cite{pfeiffer2020adapter1} by inserting an adapter module subsequent to the MLP block at each layer with a reduction factor of 8.


\begin{figure*}[t!]
    \begin{minipage}{\textwidth}
        \centering
        \includegraphics[width=0.95\textwidth]{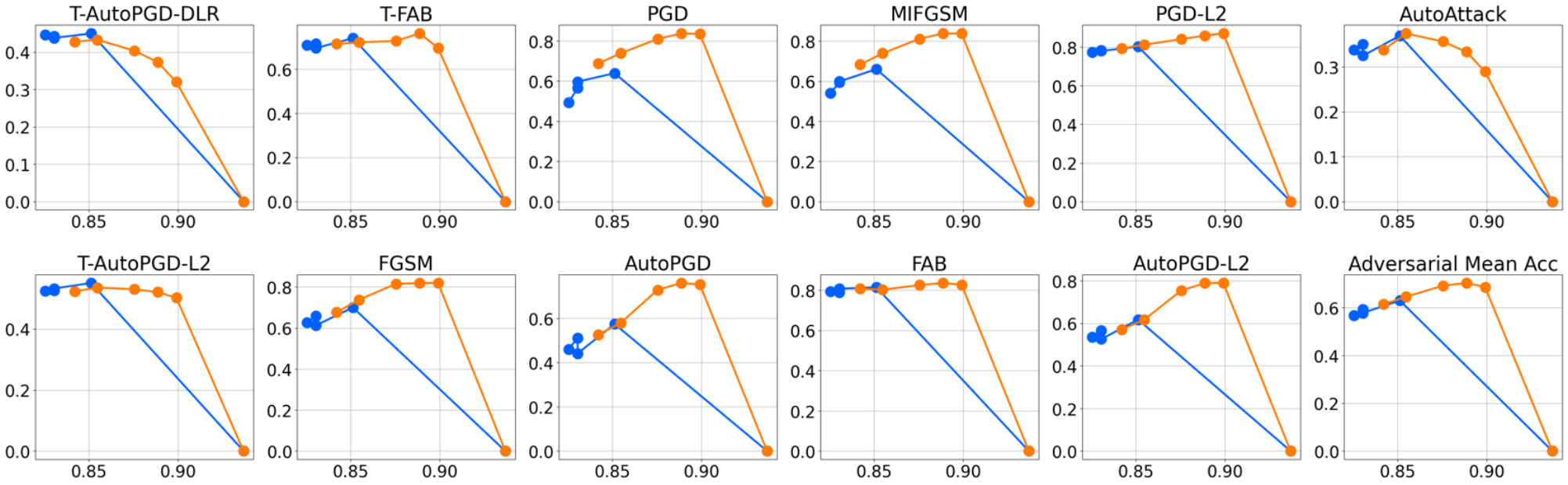}
        \subcaption{ResNet34}
        \label{fig:cifar10_adver_res34}
    \end{minipage}%
    \hfill
    \begin{minipage}{\textwidth}
        \centering
        \includegraphics[width=0.95\textwidth]{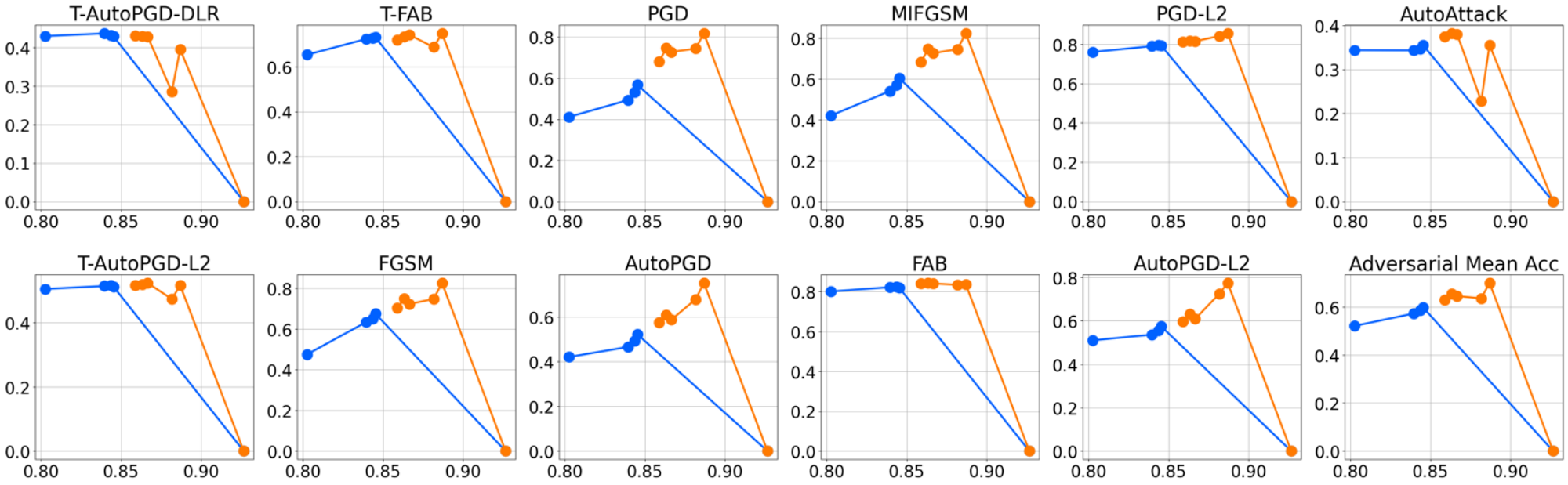}
        \subcaption{WRN-28-10}
        \label{fig:cifar10_adver_wrn}
    \end{minipage}
    \caption{SA-AA Fronts for Adversarial Training from Scratch on CIFAR10 with ResNet34 and WRN-28-10.}
    \label{fig:app:cifar10_adver}
\end{figure*}

\begin{figure*}[t!]
    \centering
    \hfill
    \begin{minipage}{\textwidth}
        \centering
        \includegraphics[width=0.95\textwidth]{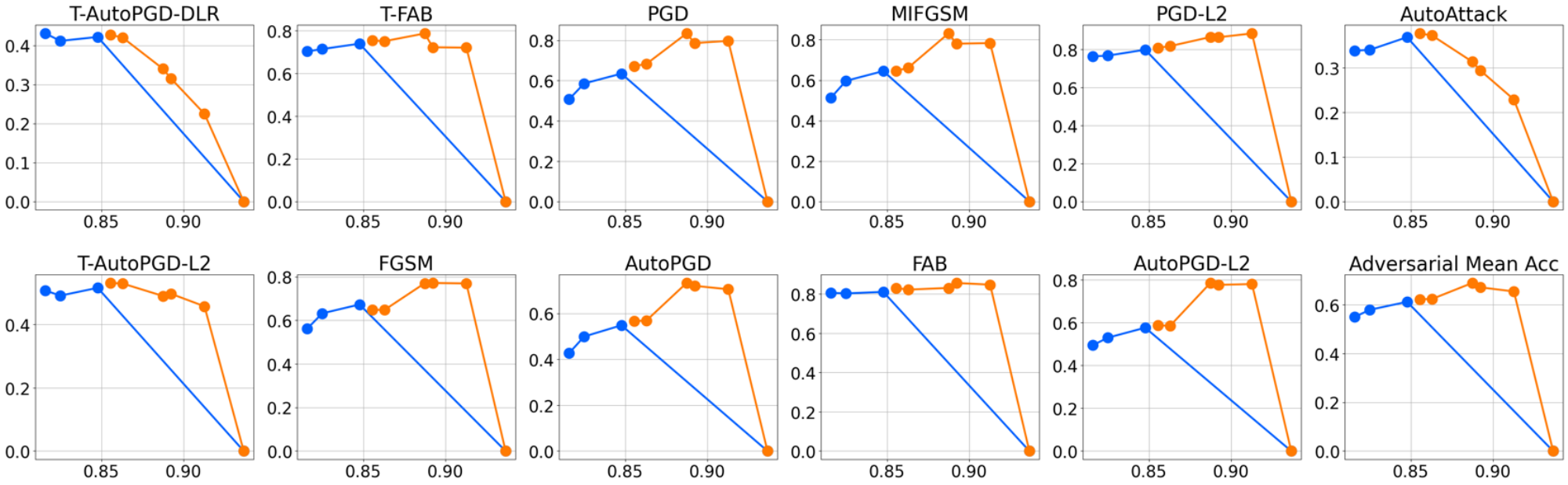}
        \subcaption{ResNet34}
        \label{fig:app:cifar10_trades_res34}
    \end{minipage}%
    \hfill
    \begin{minipage}{\textwidth}
        \centering
        \includegraphics[width=0.95\textwidth]{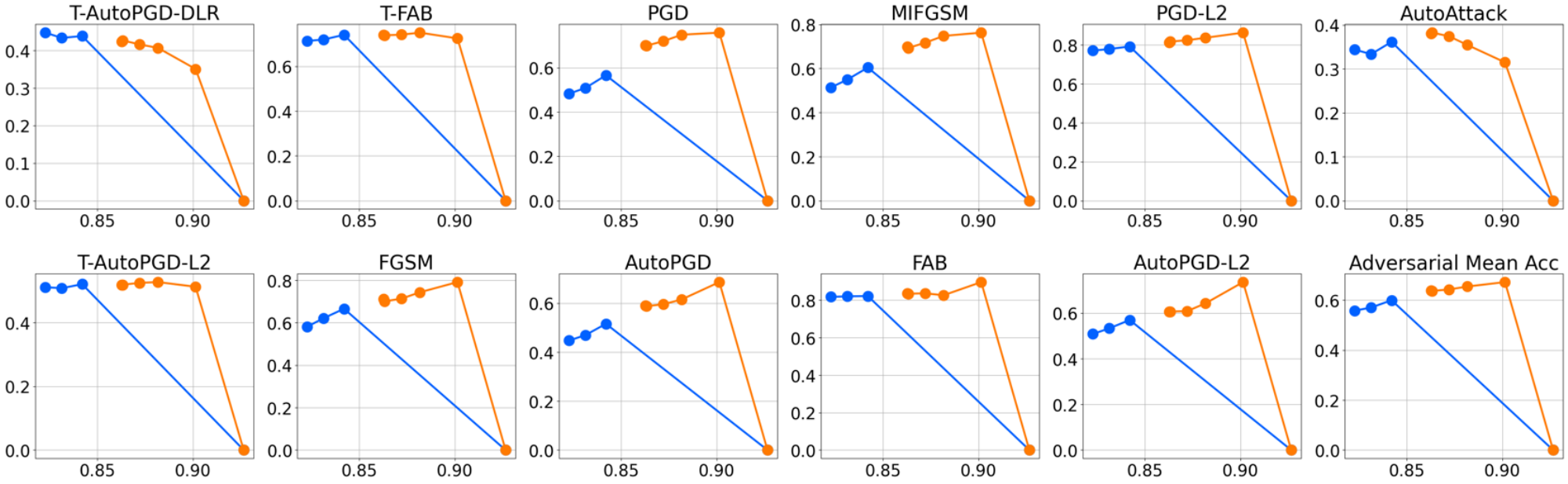}
        \subcaption{WRN-28-10}
        \label{fig:app:cifar10_trades_wrn}
    \end{minipage}
    \caption{SA-AA Fronts on CIFAR10 for Adversarial Training from Scratch using TRADES with ResNet34 and WRN-28-10.}
    \label{fig:app:cifar10_trades}
\end{figure*}

\begin{figure*}
    \centering
    \begin{minipage}{\textwidth}
        \centering
        \includegraphics[width=0.95\textwidth]{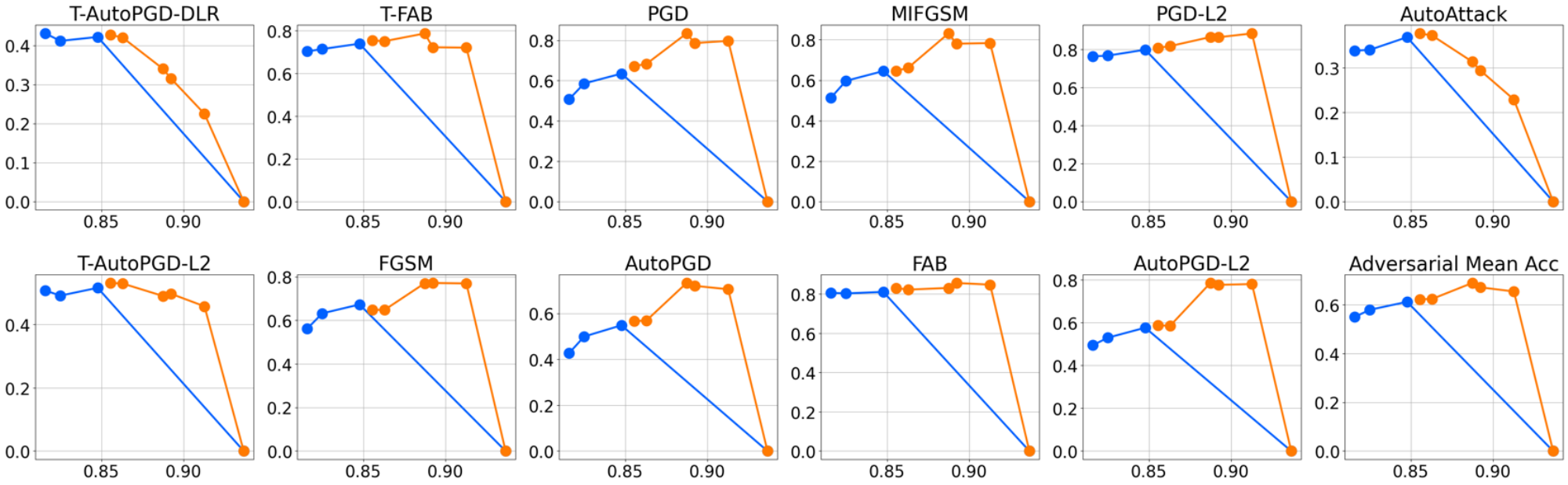}
        \subcaption{ResNet34}
        \label{fig:cifar10_clp_res34}
    \end{minipage}%
    \hfill
    \begin{minipage}{\textwidth}
        \centering
        \includegraphics[width=0.95\textwidth]{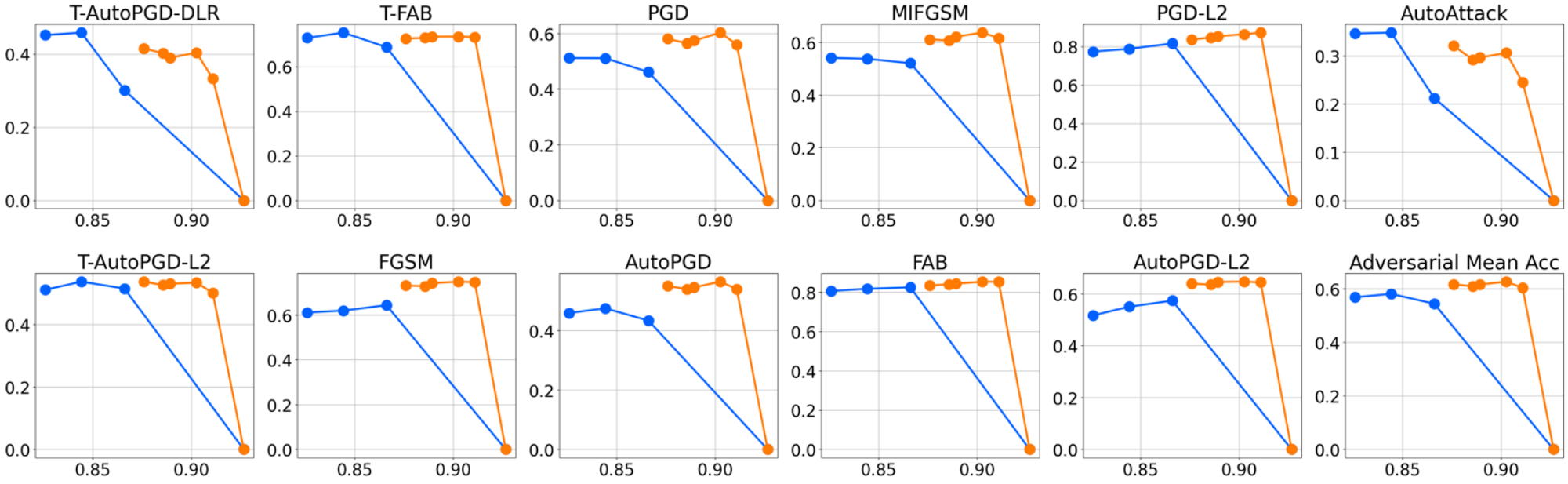}
        \subcaption{WRN-28-10}
        \label{fig:cifar10_clp_wrn}
    \end{minipage}
    \caption{SA-AA Fronts for Adversarial Training from Scratch on CIFAR10 using CLP with ResNet34 and WRN-28-10.}
    \label{fig:app:cifar10_clp}
\end{figure*}

\textbf{The effect of inner maximization solver in AT. }In \cref{tab:dif_tr_attck}, we conduct the ablation study for using different attack methods to generate adversarial samples during adversarial training from scratch. We find that PGD-DLR can achieve higher adversarial accuracies when $\gamma=0.9$ but lead them worse when $\gamma=0.8$ but not significant. We conclude that the effect of the inner maximization solver, as well as the adversarial attack method during AT, does not dominate the performance of CA-AT. However, choosing a better inner maximization solver can still help adversarial robustness marginally.

\textbf{Results for different model architectures. }For different model architectures such as ResNet34 and WRN-28-10, their SA-AA front on CIFAR10 and and CIFAR100 with different adversarial loss functions are shown in \cref{fig:app:cifar10_adver}, \cref{fig:app:cifar10_clp}, and \cref{fig:app:cifar10_trades}. All of those figures demonstrate CA-AT can consistently surpass Vanilla AT across different model architectures.

\textbf{Results for $L_{2}$-based adversarial attacks with different budgets. }Besides evaluating the adversarial accuray on $L_{\infty}$-based attacks with different budgets~(\cref{tab:dif_bounds_linf}), we also evaluate the adversarial robustness against $L_{2}$-based adversarial attacks with different budgets~($||\epsilon||_2 = [0.5,1,1.5,2]$), which is shown in \cref{tab:dif_bounds_l2}.

\textbf{Results for Black-box Attack. }To demonstrate the performance gain of adversarial accuracy is not brought by obfuscated gradients~\cite{athalye2018obfuscated}, \cref{tab:Black-Box} shows that CA-AT~($\gamma=0.8$ and $\gamma=0.9$) outperforms Vanilla AT~($\lambda=0.5$ and $\lambda=1$) against the black-box attacks such as Square~\cite{}.
\begin{table}[t]
\centering
\resizebox{\textwidth}{!}{%
\begin{tabular}{c|c|cc|cc|cc}
\hline
\multicolumn{1}{l|}{}     & \multicolumn{1}{l|}{} & \multicolumn{2}{c|}{PGD-L2}     & \multicolumn{2}{c|}{AutoPGD-L2} & \multicolumn{2}{c}{T-AutoPGD-L2}  \\ \hline
\multicolumn{1}{l|}{}     & $p=2$                 & $\gamma=0.8$    & $\lambda=0.5$ & $\gamma=0.8$    & $\lambda=0.5$ & $\gamma=0.8$    & $\lambda=0.5$   \\ \hline
\multirow{4}{*}{ResNet18} & 0.5                   & \textbf{0.8211} & 0.7759        & \textbf{0.67}   & 0.5222        & \textbf{0.5356} & 0.5327          \\
                          & 1                     & \textbf{0.8207} & 0.7748        & \textbf{0.603}  & 0.3036        & 0.261           & \textbf{0.2762} \\
                          & 1.5                   & \textbf{0.8194} & 0.7738        & \textbf{0.5652} & 0.2405        & \textbf{0.1483} & 0.1428          \\
                          & 2                     & \textbf{0.8187} & 0.7734        & \textbf{0.5331} & 0.2115        & 0.0904          & \textbf{0.088}  \\ \hline
\multicolumn{1}{l|}{}     & \multicolumn{1}{l|}{} & \multicolumn{2}{c|}{PGD-L2}     & \multicolumn{2}{c|}{AutoPGD-L2} & \multicolumn{2}{c}{T-AutoPGD-L2}  \\ \hline
                          & $p=2$                 & $\gamma=0.8$    & $\lambda=0.5$ & $\gamma=0.8$    & $\lambda=0.5$ & $\gamma=0.8$    & $\lambda=0.5$   \\ \hline
\multirow{4}{*}{ResNet34} & 0.5                   & \textbf{0.8411} & 0.78          & \textbf{0.7534} & 0.5255        & \textbf{0.5301} & \textbf{0.5325} \\
                          & 1                     & \textbf{0.8412} & 0.7791        & \textbf{0.7249} & 0.3806        & 0.2571          & \textbf{0.2683} \\
                          & 1.5                   & \textbf{0.8403} & 0.7784        & \textbf{0.7022} & 0.3462        & \textbf{0.1446} & 0.1438          \\
                          & 2                     & \textbf{0.8386} & 0.7781        & \textbf{0.679}  & 0.3196        & 0.0899          & \textbf{0.0905} \\ \hline
\end{tabular}%
}
\caption{Evaluation Results for CA-AT~($\gamma=0.8$) and vanilla AT~($\lambda=0.5$) across different $L_{2}$-based attacks with various restriction $\theta$.}
\label{tab:dif_bounds_l2}
\end{table}

\begin{table}[]
\centering
\resizebox{\textwidth}{!}{%
\begin{tabular}{c|cc|cc}
\hline
\multicolumn{1}{l|}{}           & \multicolumn{2}{c|}{ResNet18}     & \multicolumn{2}{c}{ResNet34}      \\ \hline
\multicolumn{1}{l|}{}           & Standard             & Square          & Standard             & Square          \\ \hline
Standard Training, $\lambda=0$  & \textbf{0.8824} & 0.703           & \textbf{0.8793} & 0.6893          \\
Vanilla AT, $\lambda=0.5$                   & 0.7627          & 0.6808          & 0.7768          & 0.6992          \\
Vanilla AT, $\lambda=1$                     & 0.7456          & 0.6684          & 0.7791          & 0.7064          \\
CA-AT, $\gamma=0.8$              & 0.8117          & 0.7344          & 0.8223          & \textbf{0.7476} \\
CA-AT, $\gamma=0.9$              & 0.853           & \textbf{0.7509} & 0.8445          & 0.7469          \\ \hline
\end{tabular}%
}
\caption{The results of Square attack between Vanilla AT and CA-AT with different hyperparameters~($\lambda$,$\gamma$) on CIFAR10.}
\label{tab:Black-Box}
\end{table}